\documentclass[review]{elsarticle}

\usepackage{lineno,hyperref}
\modulolinenumbers[5]

\journal{Journal of Knowledge-Based Systems}

\usepackage[figuresleft]{rotating}
\usepackage{graphicx}
\usepackage{amssymb}
\usepackage{booktabs}
\usepackage{algorithm}  
\usepackage{algorithmic}
\usepackage{amsmath}
\usepackage{multirow}
\usepackage{hyperref}
\usepackage{epstopdf}
\usepackage{subfigure}
\usepackage{color}
\usepackage{diagbox}
\usepackage{setspace}
\modulolinenumbers[1]
\usepackage[table]{xcolor}
\usepackage{tikz}
\usepackage{pgfplots}
\usepackage{soul}
\usepackage{makecell}
\usetikzlibrary{plotmarks,patterns,matrix,decorations.pathreplacing}
\pgfplotsset{compat=newest}

\definecolor{mygreen}{RGB}{0,135,54}
\definecolor{myblue}{RGB}{33,113,181}
\definecolor{myred}{RGB}{201,0,31}
\definecolor{mypurple}{RGB}{92,58,150}

\begin{document}
\begin{sloppypar}

\begin{frontmatter}

\title{Few-shot Online Anomaly Detection and Segmentation }
% \tnotetext[mytitlenote]{Fully documented templates are available in the elsarticle package on \href{http://www.ctan.org/tex-archive/macros/latex/contrib/elsarticle}{CTAN}.

\author[mymainaddress,mynewaddress]{Shenxing Wei}
\author[mymainaddress]{Xing Wei\corref{mycorrespondingauthor}}
\ead{weixing@mail.xjtu.edu.cn}
% \author[mymainaddress]{Xing Wei}
\author[mysecondaryaddress]{Zhiheng Ma}
\author[mythirdaddress]{Songlin Dong}
\author[mymainaddress]{Shaochen Zhang}

\author[mymainaddress,mythirdaddress]{Yihong Gong}

\cortext[mycorrespondingauthor]{Corresponding author.}

\address[mymainaddress]{School of Software Engineering, Xi'an Jiaotong University}
\address[mynewaddress]{Department of Computing, The Hong Kong Polytechnic University}
%\address[mymainaddress]{Faculty of Electronic and Information Engineering, Xi'an Jiaotong University}
\address[mysecondaryaddress]{Shenzhen Institute of Advanced Technology,Chinese Academy of Sciences}
\address[mythirdaddress]{Colloge of Artificial Intelligence, Xi'an Jiaotong University}

\begin{abstract}
Detecting anomaly patterns from images is a crucial artificial intelligence technique in industrial applications. Recent research in this domain has emphasized the necessity of a large volume of training data, overlooking the practical scenario where, post-deployment of the model, unlabeled data containing both normal and abnormal samples can be utilized to enhance the performance. Consequently, this paper focuses on addressing the challenging yet practical few-shot online anomaly detection and segmentation (FOADS) task. Under the FOADS framework, models are trained on a few-shot normal dataset, followed by inspection and improvement of their capabilities by leveraging unlabeled streaming data containing both normal and abnormal samples simultaneously. {Since the data stream has no ground truth labels, the model needs to filter the anomalous data by detection to avoid contaminating its parameters, which presents a significant challenge when initial samples are insufficient.} To tackle this issue, we propose modeling the feature distribution of normal images using a Neural Gas network, which offers the flexibility to adapt the topology structure to identify outliers in the data flow. In order to achieve improved performance with limited training samples, we employ multi-scale feature embedding extracted from a CNN pre-trained on ImageNet to obtain a robust representation. Furthermore, we introduce an algorithm that can incrementally update parameters without the need to store previous samples. Comprehensive experimental results demonstrate that our method can achieve substantial performance under the FOADS setting, while ensuring that the time complexity remains within an acceptable range on MVTec AD and BTAD datasets. Code is available at \url{https://github.com/Whishing/K-NG}
\end{abstract}

\begin{keyword}
Anomaly Detection, Online Learning, Neural Gas.
\end{keyword}

\end{frontmatter}

% \linenumbers

\section{Introduction}
Anomaly detection and segmentation involve the examination of images to identify rare or distinct components and to localize specific regions. In real-world assembly lines, it is essential to identify products with anomalous parts in order to ensure quality. While humans possess a natural ability to detect novel patterns in images and can easily discern expected variances and outliers with only a limited number of normal examples, the repetitive and monotonous nature of this task makes it challenging for humans to perform continuously. Consequently, there is a growing need for computer vision algorithms to assume the role of humans on assembly lines.
In modern industrial settings, the collection of normal samples is relatively straightforward, framing this task as an out-of-distribution problem where the objective is to identify outliers based on the training data distribution. Notably, industrial surface defect detection presents a unique challenge, as it requires the localization of regions containing subtle changes and more significant structural defects, rendering the problem considerably more complex.

Recent years have witnessed the development of anomaly detection and segmentation, particularly with the advent of deep learning. Early methods in the anomaly detection area primarily employ Auto-encoder~\cite{gong2019memorizing,zhang2023destseg,bergmann2020uninformed,liu2023fair} or GAN~\cite{liang2023omni,yamada2022reconstructed,zavrtanik2021draem} to learn the normal distribution. Given a test image, these works try to reconstruct the image and compare the reconstructed images with the originals. Since the model is only trained on normal data, the anomaly parts are expected to be reconstructed poorly. {Cohen \emph{et al.} ~\cite{spade} and Defard \emph{et al.}~\cite{defard2020padim} proposed the use of deep convolution neural network that trained on ImageNet dataset as feature extractor.} Although the model is pre-trained on a general image classification dataset, it still offers strong performance on anomaly detection and segmentation even without any specific adaptation. In Table \ref{sota_table}, we give a summary of part of the conventional methods.

\begin{center}
\begin{figure}[t]
    \centering
    \includegraphics[width=0.9\textwidth]{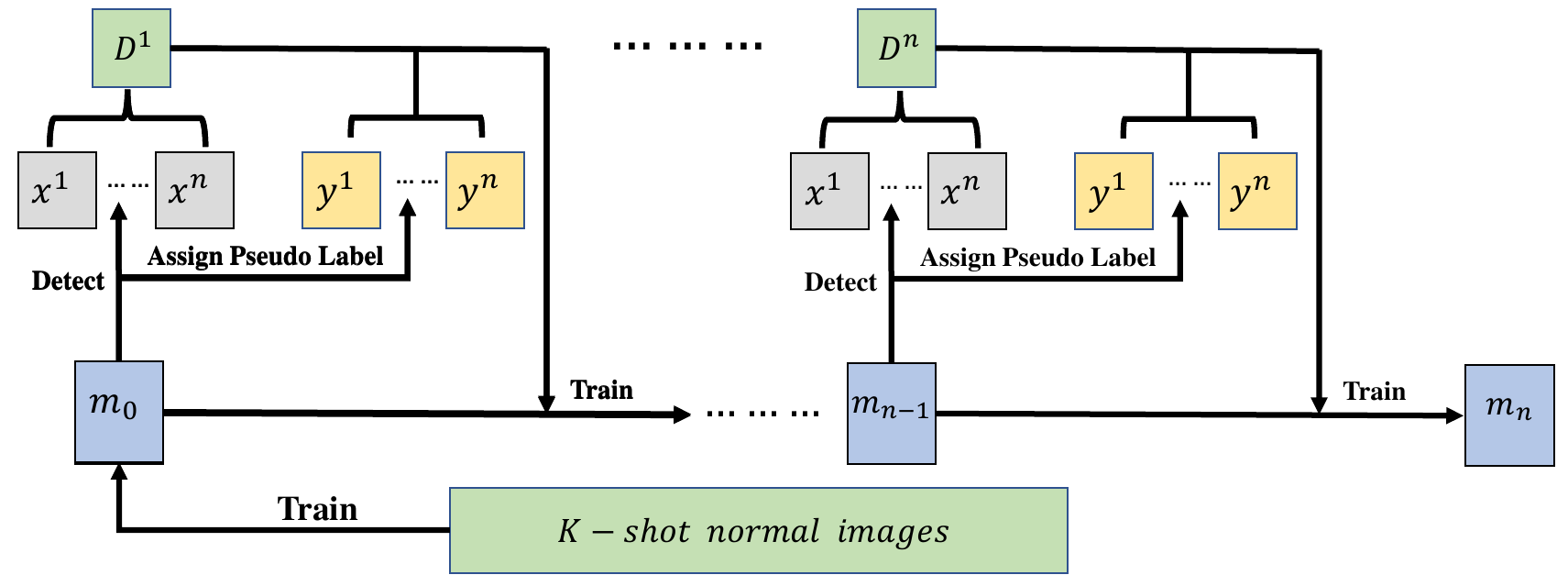}
    \centering
    \caption{Illustration of the FOADS. The model $m_0$ is built on a training set which only contains  $K$ normal images firstly. Subsequently, the model detects the test images $D^1,D^2,...,D^n$ and updates itself by  incorporating the unlabeled streaming data which contains both normal and abnormal images. Throughout the online learning process, our model assigns a pseudo label for each data and then trains itself with these labels.}
    \label{def}
    \end{figure}
\end{center}
However, there are still some unsolved limitations in these works. Advanced methods~\cite{zavrtanik2021draem, defard2020padim, roth2021towards, zhang2023diffusionad} for unsupervised industrial image anomaly detection and segmentation ignore the continuous data flow comprising a mixture of normal and abnormal data, generated in industrial scenarios after the model is deployed. Another limitation is that most current approaches need to collect a substantial number of normal images for training. From an industrial perspective, there is a reluctance to allocate significant time to collecting extensive training data. Thus, if the model can be established rapidly with limited initial training samples and constantly boosts its abilities to achieve better detection results, the practicality of the entire algorithm would be significantly enhanced.
We term this ability as few-shot online anomaly detection and segmentation (FOADS). The flowchart in Figure \ref{def} depicts our proposed framework for detecting and updating by leveraging unlabelled data which contains both normal and abnormal instances. 

{However, due to the initial training samples are insufficient, the model needs to be modified frequently and significantly during the online learning. As mentioned above, existing works in anomaly detection and segmentation primarily leverage deep reconstruction models or memory banks. Deep neural network models have been proven to tend to overfit current data and lose knowledge learned before, which is called ``catastrophic forgetting" ~\cite{wei2023topology}. Memory bank-based approaches utilize cluster algorithms such as multivariate Gaussian model and coreset-subsampling to represent the entire feature space.  They are characterized by a predetermined isolated structure, making it difficult to adapt to unlabeled mixed data flow.} 
Hence, the current anomaly detection and segmentation frameworks are not well-suited for directly addressing the FOADS problem.
In the realm of online cluster learning, many researchers propose to leverage the Neural Gas (NG) \cite{martinetz1991neural} model, which can adjust its neighborhood size or relation according to changing scenes flexibly for better adaptation. 
{Consequently, we choose NG as the base model for addressing FOADS problem.}
But existing related work utilizing NG mainly concentrates on handcraft feature descriptors~\cite{yuan2014online} 
which are hard to design and less representative than multi-scale deep features whose performance has been demonstrated in previous studies~\cite{spade,defard2020padim,roth2021towards}. Therefore, the intuitive idea is to introduce multi-scale deep features into previous NG networks to enhance the representational ability. However, the optimization algorithms of NG, such as Competitive Hebbian learning and Kohonen adjustment, are computationally intensive. The inherent limitation of these learning strategies, which update cluster centroid vectors one by one, results in an unacceptable running speed, severely limiting its practical value.

\begin{table}[!ht]\footnotesize
\centering
 \caption{A summary of conventional anomaly detection and segmentation methods }
 \renewcommand{\arraystretch}{2.0}
 \setlength\tabcolsep{2pt}
 \begin{tabular}{@{}lll@{}}
 \toprule
  \bfseries Method &\bfseries Highlights &\bfseries Disadvantages\\
 \midrule
 %  \bfseries MemAE \cite{gong2019memorizing}
 % & \makecell[l]{Proposing a memory-augmented \\ autoencoder for anomaly detection} & 
 
 \bfseries DiffusionAD~\cite{zhang2023diffusionad}
 & \makecell[l]{Leveraging diffusion model \\ for better reconstruction.} & \multirow{2}*{\makecell[tl]{Often yielding good \\ reconstruction results \\ for anomalous images\\ and suffering from\\ ``catastrophic forgetting" \\ in online learning}}\\\\
 \bfseries  DRAEM~\cite{zavrtanik2021draem} & \makecell[l]{Designing a novel method \\ to generate abnormal images}  &  \\
 \hline
 
 \bfseries  SPADE \cite{spade}&      \makecell[l]{Utilizing multi-scale features and KNN \\ for anomaly detection.}  &  \multirow{3}*{\makecell[cl]{Fixed isolated structure  \\ limits online learning \\ performance}}\\
 \bfseries  PaDim \cite{defard2020padim}&     \makecell[l]{Introducing a probabilistic representation \\ of the normal images}    &      \\
 \bfseries  PatchCore\cite{roth2021towards} &   \makecell[l]{Introducing a coreset sampling method \\ for constructing memory bank } & \\
 
 \bottomrule
 \end{tabular}
   \label{sota_table}
\end{table}

In this paper, we introduce a novel method as an effective remedy for the aforementioned challenges.
Given the initial training images, different from previous works in the anomaly detection area, we not only use feature embedding to characterize the manifold but also take the relation among the feature embeddings into account. Similar to the approach in Neural Gas, edges are added between similar feature embedding and finally, constitute a graph structure. This graph is then utilized to represent the topology of the manifold. 
To effectively distinguish between normal and abnormal data, we propose a novel threshold that adjusts adaptively based on the topology structure of the neural network. This flexible thresholding mechanism enables us to effectively screen out abnormal data in the presence of mixed normal and abnormal data flows. To reduce the computation cost of conventional NG, we employ its flexible topology structure and forego its computationally intensive components such as Hebbian learning and Kohonen algorithm. Specifically, we employ the $K$-means algorithm to expedite the learning process in this paper. Besides, considering the consumption of the memory and further improvement of the running speed, we propose to model the embedding distribution by parameter matrix and update it iteratively, eliminating the need to store the old images that have already passed through the model.

The experiments on the diverse MVTec AD and BTAD anomaly detection dataset demonstrate that our method can achieve a satisfying result in solving FOADS problem. With a competitive computation speed, our method can attain a gradually rising accuracy in the procedure of processing mixing data flow. Our contributions can be summarized as followings:
\begin{itemize}
  \item [1)] 
  We recognize the importance of few-shot online anomaly detection and segmentation (FOADS) and define a problem setting to better organize the FOADS research. Compared with the widely studied anomaly detection and segmentation, FOADS is more challenging but more practical.  
  \item [2)]
   We propose a flexible framework based on Neural Gas, termed $K-$NG, to address the few-shot image anomaly detection and segmentation problem in online scenarios. We also propose a series of algorithms to ensure the speed and memory consumption are acceptable.
  \item [3)]
  Our method is evaluated on MVTec AD and BTAD datasets which are specially designed for real-world workpieces and yields significant results. The detection and segmentation accuracy is improved greatly when data flows in continuously. The results also outperform previous methods in anomaly detection area.
\end{itemize}

The rest of this paper is organized as follows: Section \ref{relatedwork} reviews related works. Section \ref{approach} describes our algorithms. Section \ref{experiments} conducts the experiments. Section \ref{dis} is a set of the discussions, and Section \ref{conclusions} is the conclusion.

\section{Related Work}
\label{relatedwork}
\subsection{Anomaly Detection and Segmentation}
Anomaly detection and segmentation are two important research topics that try to detect novel image and pixel after observing normal samples only. To tackle the challenge that no anomaly sample is available during training, a series of approaches have been proposed. Many existing works leveraging deep learning focus on generative models including Auto-encoder \cite{gong2019memorizing,zhang2023destseg,bergmann2020uninformed,liu2023fair, zhang2023diffusionad} or Generative Adversarial Networks \cite{liang2023omni,yamada2022reconstructed,zavrtanik2021draem}. The main idea of these approaches is training the networks on the normal data by minimizing the image reconstruction loss. The anomalous region can be detected by comparing the test image and the reconstructed one pixel by pixel since the model is only trained to rebuild normal images.

Since it is hard to only leverage normal images for building deep neural networks, another group of researchers try to synthesize defects to replace the real anomaly during the training phase. For instance, Li \emph{et al.} \cite{li2021cutpaste} propose to apply a data augmentation strategy called ``CutPaste" to generate anomalous regions in the image. Zavrtanik \emph{et al.} \cite{zavrtanik2021draem} introduce an outer natural dataset for better synthesis. They randomly select regions from normal images and mix the selected parts with other natural pictures. To improve recognition accuracy, they also train a discriminate network to replace the Mean Squared Error (MSE) criterion which always leads to large deviations.

However, the significant reconstruction error due to the diverse details in the image  makes the accuracy considerably unsatisfactory. To address this issue, other researchers propose to model the normal features extracted by a CNN pre-trained on the ImageNet with a memory bank \cite{spade,roth2021towards} or probability density function \cite{defard2020padim,yi2020patch}. Then the anomaly region can be detected by measuring the distance between test image and normal distribution.  The Mahalanobis distance is often employed  for measuring the distance between the test data and normal distribution, but the normal distribution does not always obey the Gaussian distribution which is the hypothesis of Mahalanobis distance. Therefore, some researchers suggest training a ``Normalizing Flow" network to transform normal features into a standard normal distribution and give the anomaly score by deviation from standard mean \cite{yu2021fastflow,gudovskiy2021cflow}.

In contrast to these anomaly detection and segmentation works, we focus more on a practical application scenario where the number of training samples is limited, but unlabeled data flow is available.

\subsection{Online Learning and Detection}
Online learning aims at handling streaming data, allowing the learning process to continue as data is collected \cite{de2019continual,fini2020online,vs2023towards}.  A typical application of online learning is the detection in changing scenes, especially in video object detection. Kang \emph{et al.} \cite{kang2016object} propose a CNN based framework that leverages simple object tracking for video object detection. But this method consists of several different stages, resulting in high time complexity in video detection. Recently, Wu \emph{et al.}~\cite{Wu_2023_ICCV} propose a plug-and-play module for label-efficient framework for online object detection from streaming video.

In the realm of online anomaly detection, existing work primarily concentrates on video processing area \cite{doshi2020continual,10447554}. In order to enable the model to adjust its structure with the data flow, some previous work proposed an online Self-Organizing Map (SOM) model and obtained some certain results. Nevertheless, the SOM has an inherent limitation that lattice structure is fixed which makes some null nodes far away from realistic samples. To address this limitation, some researchers propose a more flexible network called Neural Gas which learns data topology utilizing Hebbian learning.

In this paper, considering the limited train set can be inadequate, which makes the model need to be modified frequently during online learning process, we also employ the flex structure of Neural Gas. To mitigate the problem that Hebbian learning is too slow, we replace it with the K-means algorithm, which converges rapidly.

\subsection{Few-shot Learning}
Few-shot learning aims to enable the model to recognize unseen novel patterns while just leveraging very few training samples. To achieve few-shot learning, researchers usually adopt techniques like metric learning or optimization-based algorithm \cite{cheng2023frequency,song2023comprehensive,ganea2021incremental}.  Optimization-based methods often employ a meta-learner which can generate weights for a new learner to process new tasks.  In order to generate the meta-learner, researchers generally apply a separate network \cite{zhang2024metadiff} to produce learners or model it as an optimization procedure \cite{sun2024meta}.  Methods based on metric learning attempt to learn a feature embedding space where the same class is close while different classes are far apart. To build the distinguishable embedding space, Shao \emph{et al.} \cite{shao2021few} uses a Siamese network which can minimize the distance between the same class and maximize otherwise. Then prototypical networks that compute per-class representation and relation networks \cite{huang2023sapenet} which learn both an embedding and a distance function are proposed.

Some researchers extend few-shot learning to object detection and instance segmentation. Kang \emph{et al.} \cite{kang2019few} generate a meta-learner based on YOLOv2 directly while Wang \emph{et al.} propose a two stage method called TFA  \cite{wang2020frustratingly} which trains Faster R-CNN and then only fine-tunes the predictor heads, achieving state-of-the-art results in object detection. In the aspect of the instance segmentation, base-metal learner \cite{lang2023base} and Siamese DETR \cite{chen2023siamese} are developed to compute embedding of the train set and combine the embedding with feature map extracted by the backbone network.

In anomaly detection and segmentation area,  some researchers have recognized the importance of training with limited samples, they try to reduce the size of the training set gradually while retaining the accuracy in an acceptable level \cite{zavrtanik2021draem,roth2021towards}.
\begin{center}
\begin{figure}[ht]
    \centering
    \includegraphics[width=0.9\textwidth]{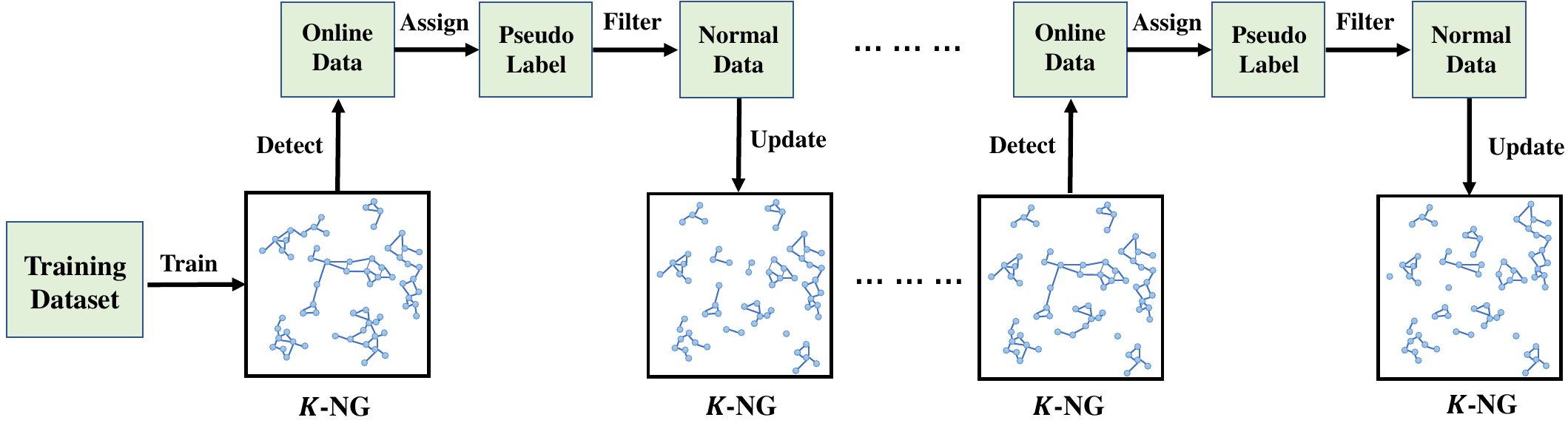}
    \centering
    \caption{The pipeline diagram of the proposed algorithm. Firstly, the initial $K$-NG model is trained on the few-shot training dataset. Subsequently, the initial model is employed to detect online data and assign pseudo labels to filter anomalous data. Finally, the remaining normal data is used to update parameters and adjust topology structures of the model.}
    \label{kngpip}
    \end{figure}
\end{center}

\subsection{Few-shot Online Anomaly Detection}
While there is a dearth of research on few-shot online anomaly detection and segmentation for industrial images, several existing literature has focused on detection of other data structures ~\cite{doshi2020continual, meng2023explainable, li2023few}. Meng \emph{et al.} ~\cite{meng2023explainable} propose leveraging a meta-knowledge learner for continuously adapting the model to new wave signal sources. Li \emph{et al.} ~\cite{li2023few} exploit a dueling triplet network to learn domain-invariant feature representation of time series and incremental update the classifier.
However, these approaches are limited to predicting whether the sample is anomalous or not, which is not suitable for industrial image anomaly detection. In this paper, we not only provide the classification label but also predict per-pixel scores for accurate locating.

\section{Few-shot Online Anomaly Detection and Segmentation}
\label{approach}
We define the few-shot online anomaly detection and segmentation as follows: 

Suppose we have $K$ ($K$ is a small integer) normal samples in training set denoted as $X=\{x_1,x_2,...x_K\}$, and a stream of unlabeled test samples
$D=\{D^{(1)},D^{(2)},...,D^{(N)}\}$ which is far larger than the training samples. The model $\Theta$ is trained on the $K$ training samples firstly. And then, the model sequentially detects and segments images that contain anomalous regions on the unlabeled set $D^{(1)}, D^{(2)},...D^{(N)}$. During the detection process, the model updates parameters by its own distinguishment results on the  incoming data stream, aiming to enhance its detection capability. The main challenges are  twofold : (1) filtering the anomalous data to avoid the model parameter being polluted; (2) establishing the model with limited training data.
% For training set $X$, we denote the setting with $K$ training samples as $K-shot$ FOADS. 

Our approach comprises several components, which we will describe sequentially: the initialization of the model, the algorithm for updating the model and the anomaly detection criterion. And we give a whole schematic of proposed system in Figure \ref{kngpip}.

\begin{center}
\begin{figure}[ht]
    \centering
    \includegraphics[width=0.9\textwidth]{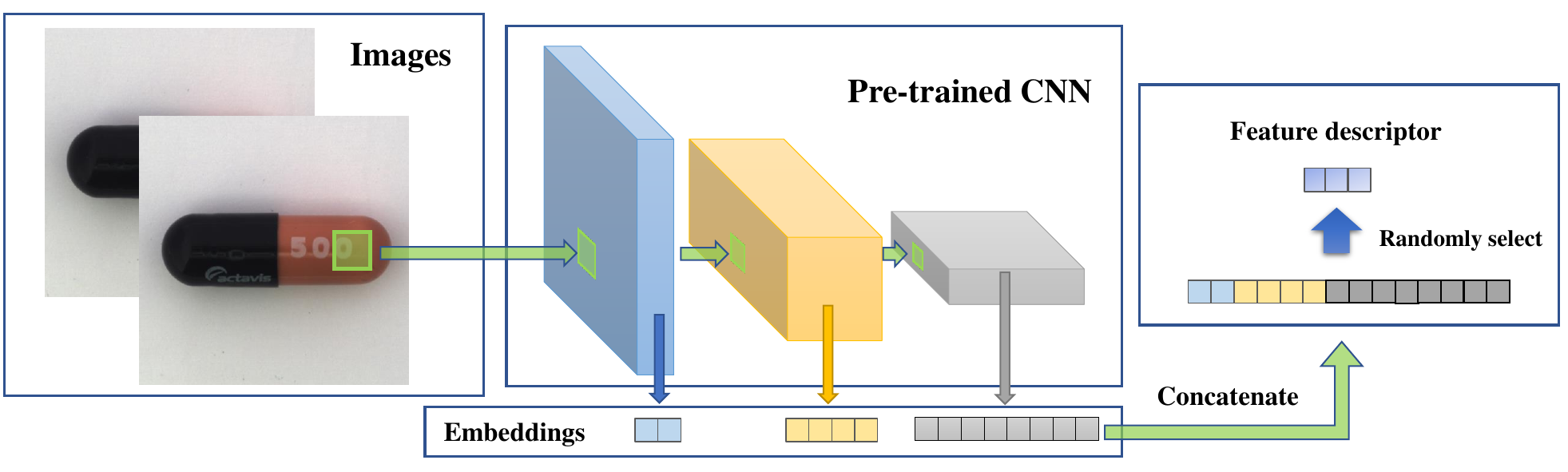}
    \centering
    \caption{The diagram illustrates the feature extraction pipeline, wherein the images are input into a pre-trained CNN to extract feature embeddings at various scales. The embeddings obtained from different CNN blocks are concatenated and then dimensionally reduced through random selection.}
    \label{fd}
    \end{figure}
\end{center}

\subsection{Feature Extraction}
Previous works utilizing Neural Gas for online clustering primarily focused on handcraft feature descriptors~\cite{yuan2014online,li2013anomaly}. In order to achieve competitive performance, we opt to utilize patch deep embeddings extracted by Convolution Neural Network (CNN) that has been pre-trained on the ImageNet dataset. Due to the presence of downsampling in CNN, each embedding in the CNN feature map is associated with a spatially corresponding patch of the image. An interpolation algorithm is then applied to obtain a high-resolution dense map

Feature maps extracted from different CNN layers contain information from different semantic levels and resolutions. To capture both fine-grained and global contexts, we concatenate feature embeddings from different layers.  {Specifically, as shown in the Table \ref{resnet18}, features at the end of the first residual block conv2\_x, the second residual block conv3\_x, and the third residual block conv4\_x from the ResNet18 backbone are adopted.} Similarly, as proposed by Defard \emph{et al.} \cite{defard2020padim},  we apply a simple random dimensionality reduction to decrease the complexity while maintaining high performance. The process schematic for the feature embedding extraction is depicted in Figure \ref{fd}.

\begin{table}[!hb]
\centering
\footnotesize
  \caption{Structure of the ResNet18 based feature embedding extractor.}
 \renewcommand{\arraystretch}{1.2}
 \centering
 \begin{tabular}{ccc}
 \toprule
 \bfseries Layer Name & \bfseries Output Size  &\bfseries Convolution Kernel  \\
 \midrule
 conv1 & 112 $\times$ 112  $\times$ 64  & 7 $\times$ 7, 64, stride 2\\
 \midrule
 conv2\_x & 56 $\times$ 56  $\times$ 64  &  \makecell{ 3 $\times$ 3 max pool, stride 2\\  $\left[\begin{array}{cc}
     3\times3,  & 64 \\
     3\times3,  & 64
        \end{array} \right] \times 2$ }   \\
\midrule
 conv3\_x & 28 $\times$ 28 $\times$128   &  $\left[\begin{array}{cc}
                                3\times3,  & 128 \\
                                3\times3,  &  128
                            \end{array} \right] \times 2$ \\
                            \midrule
 conv4\_x & 14 $\times$ 14 $\times$ 256   &  $\left[\begin{array}{cc}
                                3\times3,  & 256 \\
                                3\times3,  &  256
                            \end{array} \right] \times 2$ \\
                            \midrule
 conv5\_x & 7 $\times$ 7  $\times$ 512  &  $\left[\begin{array}{cc}
                                3\times3,  & 512 \\
                                3\times3,  &  512
                            \end{array} \right] \times 2$ \\
                            \midrule
      fc      & 1 $\times$ 1    &  average pool, 1000-d fc, softmax \\
 \bottomrule
 \end{tabular}
  \label{resnet18}
\end{table}

\begin{center}
\begin{figure}[ht]
\centering
    \subfigure[]{\includegraphics[width=5cm]{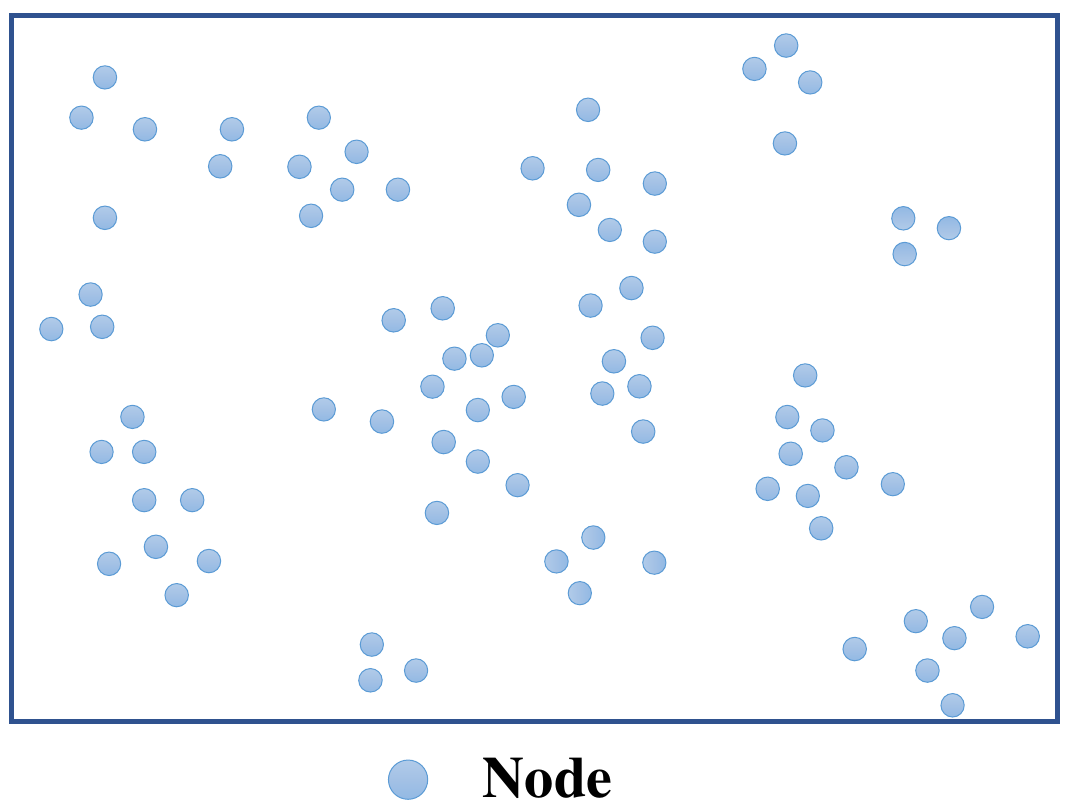}}
  \subfigure[]{\includegraphics[width=5cm]{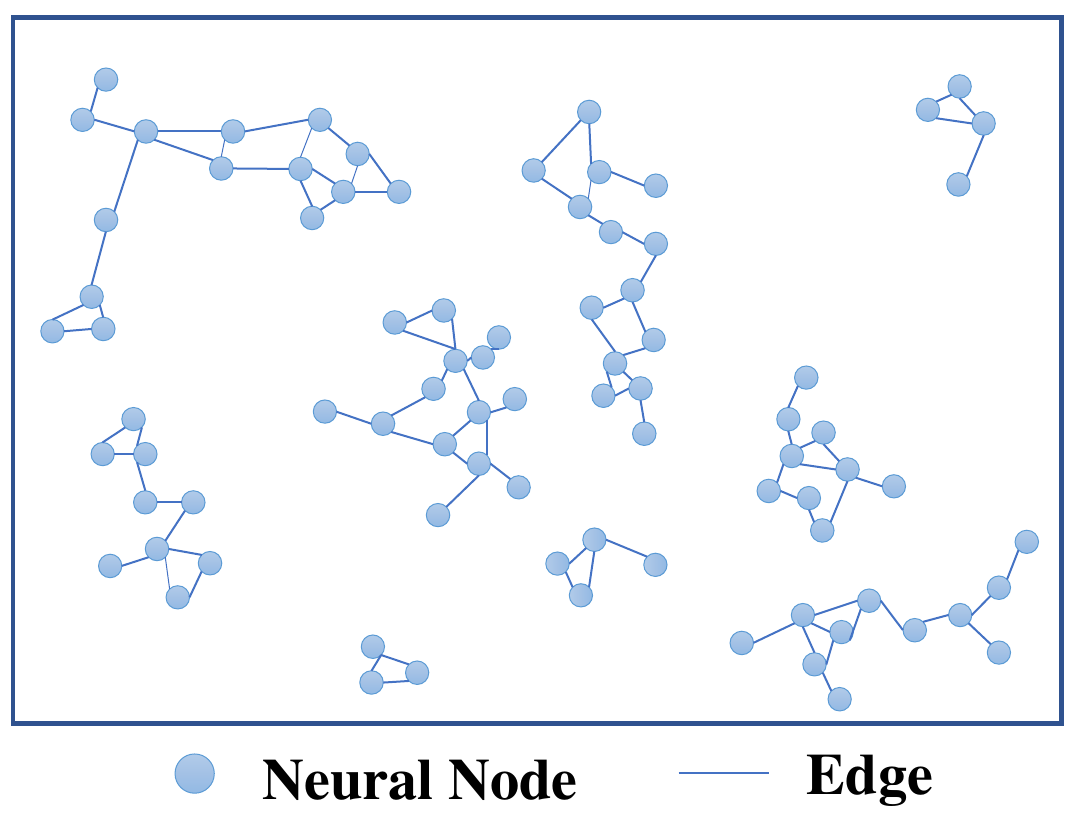}}
    \caption{ Comparison of two methods for characterizing the normal feature manifold: Figure (a) depicts the naive memory bank algorithm used to model the normal manifold, while Figure (b) illustrates the Neural Gas model, which utilizes a graph to establish topology relations among different feature descriptors. The topology structure of the Neural Gas model provides it with enhanced representational capacity. }
    \label{model}
    \end{figure}
\end{center}
\subsection{Initialization of the Neural Gas}
After a random node initialization, the original Neural Gas learns the data distribution via Competitive Hebbian learning. Given an input feature embedding $x$, the nearest neuron to it $s_1$, the new weight $w_{s_1}$ of neuron $s_1$ is updated by:
\begin{equation}
    w_{s_1}=w_{s_1}+\lambda\|w_{s_1}-x\|_2,
\end{equation}
where $\lambda$ denotes the learning rate of the network. However, as the volume of data increases, the computational time required by this learning rule becomes impractical for real-world applications. To mitigate this issue, we propose to enhance the convergence speed by integrating the cluster algorithm $K$-means with Neural Gas, resulting in the $K$-NG method.

The initial neuron nodes are generated through random selection on the training set. During the clustering process, the distance to every neuron node is calculated for each feature embedding in the training set. Subsequently,  the feature embedding is assigned to the nearest node $s_1$ for parameter updating. In order to model the relations between different neuron nodes, we employ a topology graph introduced in Neural Gas to represent the relation structure. Figure \ref{model} gives a sketch of Neural Gas and a simple comparison with other memory bank methods. 

Following this, the second nearest neuron node $s_2$ is also identified for establishing the graph: If there is no existing edge between $s_1$ and $s_2$, a new connection with a parameter termed "$age$" is added between them; This parameter is employed to describe the activity level between two nodes. If there is already an edge, the $age$ of it is reset to 0. Then the $age$ of all the other edges in the network increases by 1, and edges whose $age>age_{max}$ are removed.  Finally, we calculate the mean, threshold, and covariance matrix of the data assigned to each neuron node.  In the algorithm mentioned above, the covariance matrix of a neuron node is estimated by:
\begin{equation}
    \Sigma=\frac{1}{N-1}\sum_{k=1}^{N}(x^k-\mu)(x^k-\mu)^T+\epsilon I,
\label{cov}
\end{equation}
where the $\{x^k | k\in [1,N]\}$ are the feature embeddings assigned to the given neuron node, $N, \mu$ is the number of embeddings and sample mean respectively; the term $\epsilon I$ is the regularisation which makes the sample covariance matrix full rank and invertible.  The threshold $T_{s_i}$ is calculated by using the median distance between neuron $s_i$ and its neighboring neurons that are connected by edges:
\begin{equation}
    T_{s_i}=\underset{n \in \mathcal{N}_{s_i}}{\operatorname{median}} \| \mathcal{A}_{s_i} -  \mathcal{A}_{n} \|_2,
\end{equation}
where $\mathcal{N}_{s_i}$ is the neighbor set. If neuron $s_i$ has no neighbor, the threshold $T_{si}$ is defined by the minimum distance between $s_i$ and other neurons:
\begin{equation}
    T_{s_i}=\underset{\mathcal{A}_{n} \in \mathcal{A} \setminus \mathcal{A}_{s_i}}{\operatorname{min}} \| \mathcal{A}_{s_i} -  \mathcal{A}_{n} \|_2.
\end{equation}
Finally, the weights of each neuron node are updated with the sample mean computed from the feature embeddings assigned to it, and the entire algorithm is executed for several epochs. The complete algorithm can be summarized as \textbf{Algorithm} \ref{init}.

\begin{algorithm}[ht]
\caption{The initialization of $K-$NG network} 
\label{init}
\hspace*{0.02in} {\bf Input:} 
Data set $X=\{x_1,x_2,...,x_M\}$ \\
 \hspace*{0.02in} {\bf Input:} 
$k$ target number of cluster center\\
 \hspace*{0.02in} {\bf Require:} 
The number of training epoch $Eps$ \\
 \hspace*{0.02in} {\bf Require:} 
The maximum age of edge $age_{max}$ \\
\hspace*{0.02in} {\bf Output:}
Cluster center $\mathcal{A}=\{\mathcal{A}_1,\mathcal{A}_2,...,\mathcal{A}_k\}$\\
% \hspace*{0.02in} {\bf Output:}
% Cluster error $E=\{E_{\mathcal{A}_1},E_{\mathcal{A}_2},...,E_{\mathcal{A}_k\}}$ \\
\hspace*{0.02in} {\bf Output:}
The number of vectors $N=\{N_{\mathcal{A}_1},N_{\mathcal{A}_2},...,N_{\mathcal{A}_k\}}$ \\
\hspace*{0.02in} {\bf Output:}
The covariance matrix $\Sigma=\{\Sigma_{\mathcal{A}_1},\Sigma_{\mathcal{A}_2},...,\Sigma_{\mathcal{A}_k\}}$ \\
\begin{algorithmic}[1]
\STATE Randomly choose $k$ cluster  center from $X$ to generate initial $\mathcal{A}$
\FOR{$i$=1,2,...,$Eps$: } 
\STATE Calculate the distance matrix $D$ from each $x_l$ to each cluster center $\mathcal{A}_n$
\STATE Clear $N,\Sigma$
\FOR{$l$=1,2,...,$M$: }
\STATE $s_1=\underset{\mathcal{A}_{s} \in \mathcal{A}}{\operatorname{argmin}}\| x_l - \mathcal{A}_s\|_2 $
\STATE $s_2=\underset{\mathcal{A}_{s} \in \mathcal{A} \setminus \mathcal{A}_{s_1}}{\operatorname{argmin}}\| x_l - \mathcal{A}_s\|_2$
\STATE Add an edge between $s_1$, $s_2$ whose $age$ is 0
\STATE The $age$ of other edges in the network increase 1
\STATE Delete edges whose $age$ $>$ $age_{max}$
\STATE Assign $x_l$ to cluster center $\mathcal{A}_{s_1}$
% \STATE $E_{\mathcal{A}_{s1}}=E_{\mathcal{A}_{s1}}+\| x_l - \mathcal{A}_{s1}\|_2$
\STATE $N_{\mathcal{A}_{s_1}}=N_{\mathcal{A}_{s_1}}+1$
\ENDFOR
\STATE Recalculate means $\mathcal{A}$, covariance matrix $\Sigma$ for vectors assigned to each cluster by Equation \ref{cov}\\
\STATE Update the cluster center with the new means
\STATE Update the threshold of each cluster center
\ENDFOR
\STATE {\bf Return:} $\mathcal{A},N,\Sigma$
\end{algorithmic}
\end{algorithm}

\subsection{Online Learning of the Neural Gas}

\begin{figure}[!ht]
\centering
\subfigure[Distance between $x$ and $s_1$ is greater than threshold $T_{s_1}$]{\includegraphics[width=4cm]{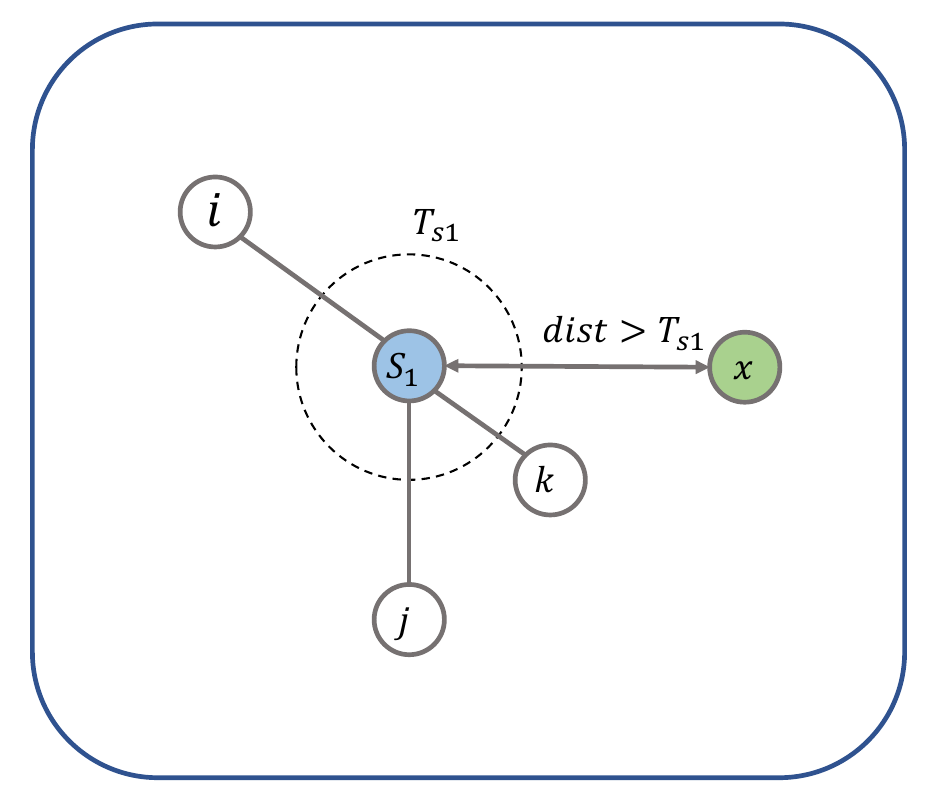}}
  \subfigure[There is no existing connection between $s_1$ and  $s_2$]{\includegraphics[width=4cm]{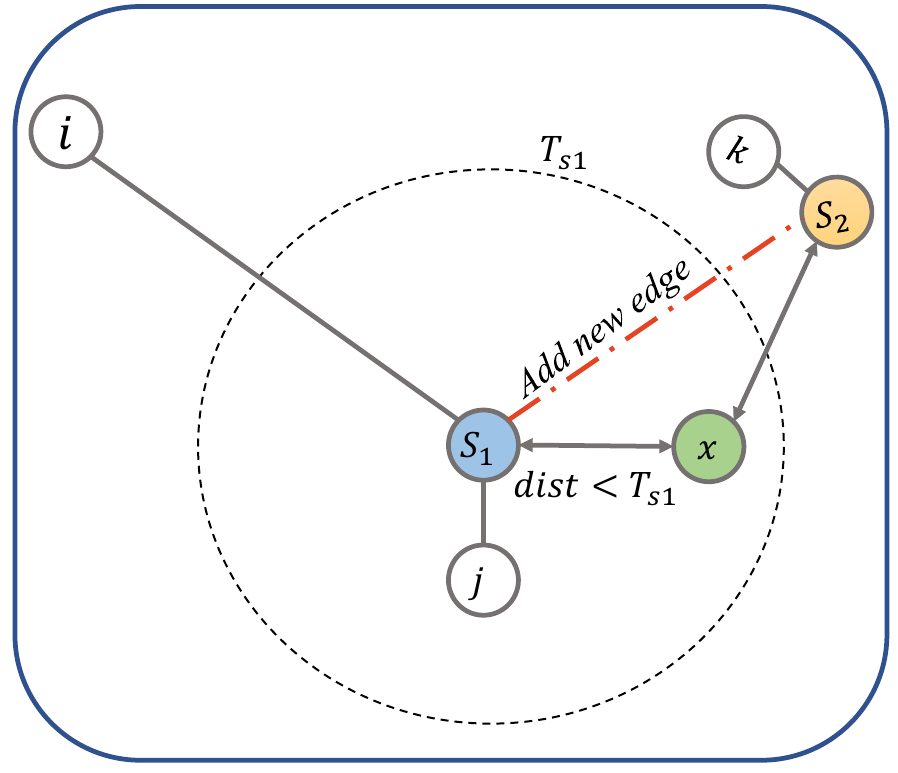}}
  \subfigure[There is existing connection between $s_1$ and  $s_2$]{\includegraphics[width=4cm]{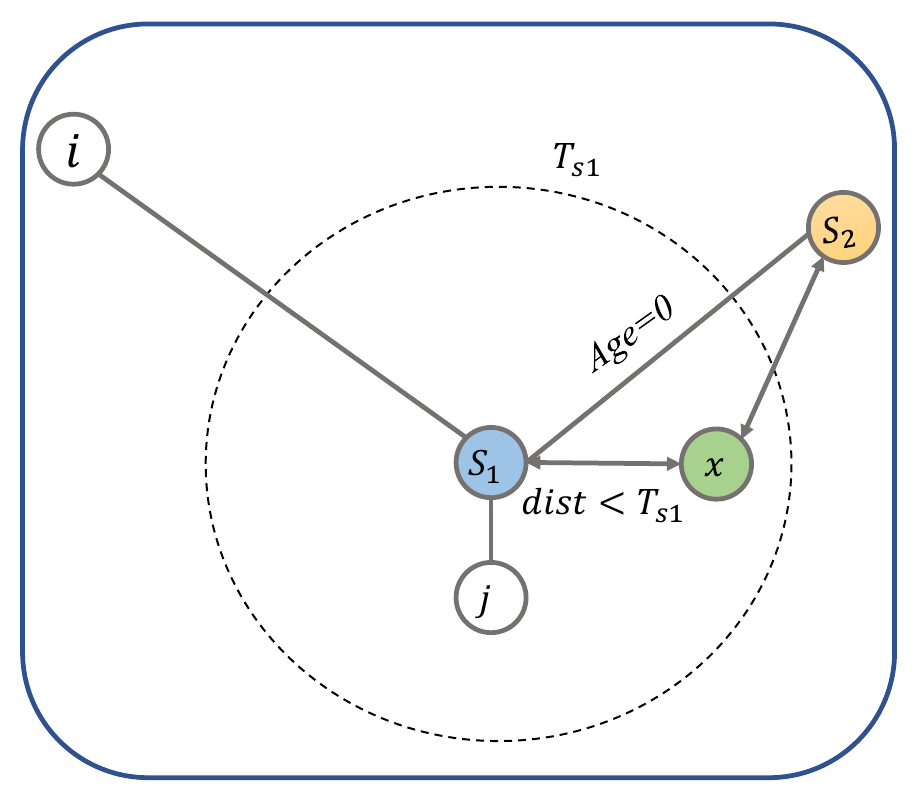}}
\caption{Diagram of the online learning process. Figure (a) shows the circumstance that the distance between $x$ and $s_1$ exceeds the threshold  $T_{s_1}$, resulting in the abandonment of $x$. In Figure (b), the distance between $x$ and $s_1$ is less than threshold $T_{s_1}$, but there is no existing connection between $s_1$ and $s_2$. Therefore, a new edge is added between them, and $x$ is assigned to $s_1$. In the situation depicted in Figure (c), where there is an existing connection between $s_1$ and $s_2$, the age of the existing edge is reset to 0   }
\label{onlineng}
\end{figure}

Upon the arrival of a new data stream batch into the model, the parameters are updated based on the characteristics of the incoming data.  In the updating algorithm, the network first assigns a pseudo label to each new feature embedding. The label is allocated based on the distance $d$ between input embedding and its nearest neuron node $s_1$. If $d$ is larger than the threshold of $s_1$, the input feature embedding is classified as anomalous data; otherwise, it is categorized as normal and assigned to $s_1$. Secondly, the relation graph in the network is updated in a similar manner as \textbf{Algorithm} \ref{init}. A diagram illustrating the online learning process is presented in Figure \ref{onlineng}.

All new feature embeddings divided into normal class by the network participate in the update procedure, while abnormal ones are abandoned. After obtaining the new data $X^{c}=\{x_1^c,x_2^c,...,x_N^c\}$ which has been assigned to $s_i$, the mean $\mu$ is computed in the first step by:
\begin{equation}
    \mu_{new}=\frac{1}{N}\sum_{k=1}^{N}x_k^c.
\label{mu}
\end{equation}

Then the covariance matrix $\Sigma_{new}$ of new data is obtained via:
\begin{equation}
    \Sigma_{new}=\frac{1}{N-1}\sum_{k=1}^{N}(x_k^c-\mu_{new})(x_k^c-\mu_{new})^T.
    \label{sigma1}
\end{equation}
The online learning algorithm incrementally updates the model parameters to conserve memory. First, the overall sample mean is computed by leveraging $N_{\mathcal{A}_i},\mathcal{A}_i$:
\begin{equation}
    \mu_{all}=\frac{1}{N_{\mathcal{A}_i}+N}(N_{\mathcal{A}_i}\mathcal{A}_i+N\mu_{new}),
    \label{mu2}
\end{equation}
Afterwards, the latest estimated covariance matrix of neuron node $s_i$ is calculated by combining $\mathcal{A}_i,\mu_{all},\mu_{new},\Sigma_{\mathcal{A}_i}, \Sigma_{new}$:

\begin{equation}
\footnotesize
\begin{split}
     &\Sigma_{\mathcal{A}_i}=\\
     &\frac{1}{N_{\mathcal{A}_i}+N-1}(N_{\mathcal{A}_i}-1)
     \Sigma_{\mathcal{A}_i}+N_{\mathcal{A}_i}(\mu_{all}-\mathcal{A}_i)(\mu_{all}-\mathcal{A}_i)^T+\\
     &\frac{1}{N_{\mathcal{A}_i}+N-1}(N-1)\Sigma_{new}+N(\mu_{all}-\mu_{new})(\mu_{all}-\mu_{new})^T,
\end{split}
\label{incre}
\end{equation}
Finally the parameter of neuron node $s_i$ is replaced by the new sample mean:
\begin{equation}
    \mathcal{A}_i=\mu_{all},
    \label{cen}
\end{equation}

In order to further demonstrate the correction of Equation \ref{incre}, we give a brief proof in appendix \ref{proof}.
And the overall algorithm flow is summarized in \textbf{Algorithm} \ref{online}.

\begin{algorithm}[!ht]
\caption{Online learning of the $K-$NG network} 
\label{online}
\hspace*{0.02in} {\bf Input:} 
Online input $X=\{x_1,x_2,...,x_M\}$ \\
 \hspace*{0.02in} {\bf Input:} 
Network parameters: $\mathcal{A},N,T,\Sigma$\\

\begin{algorithmic}[1]
\FOR{$i$=1,2,...,$M$: } 
\STATE Calculate the distance between $x_i$ and each vector in $\mathcal{A}$
\STATE $s_1=\underset{\mathcal{A}_{s} \in \mathcal{A}}{\operatorname{argmin}}\| x_i - \mathcal{A}_s\|_2 $
\STATE $s_2=\underset{\mathcal{A}_{s} \in \mathcal{A} \setminus \mathcal{A}_{s_1}}{\operatorname{argmin}}\| x_i - \mathcal{A}_s\|_2$
\IF{ $\| x_i-\mathcal{A}_{s_1} \| \leq T_{s_1}$ }
\STATE $x_i$ is assigned to the neuron $\mathcal{A}_{s_1}$
\IF{$s_1$ and $s_2$ have edge connection}
\STATE Set the age of edge between $s_1$,$s_2$ to 0
\ELSE
\STATE Add a new edge between $s_1$ and $s_2$
\ENDIF
\STATE Age of other edges in the network increase by 1
\STATE Delete edges whose $age>age_{max}$
% \STATE $E_{\mathcal{A}_{s1}}=E_{\mathcal{A}_{s1}}+\| x_i - \mathcal{A}_{s1}\|_2$
\ELSE
\vspace{0.5em}
\STATE \textbf{Continue}
\vspace{0.5em}
\ENDIF
\ENDFOR
\STATE Update the covariance matrix $\Sigma$, threshold $T$ and $\mathcal{A}$ incrementally according to equation \ref{mu},\ref{mu2},\ref{sigma1},\ref{incre},\ref{cen}.
\STATE {\bf Return:} $\mathcal{A},N,T,\Sigma$
\end{algorithmic}
\end{algorithm}

% \begin{equation}
% \begin{strip}
%     x=1
% \end{strip}
% \end{equation}
% When new batch data is put into the model, the Euclid distance between the input and the neuron is calculated firstly. Then the nearest neuron $s_1$ and second nearest neuron $s_2$ are selected by the same formula in the Algorithm \ref{init}. Given an input $x$, if 
% \begin{equation}
%     \| x-\mathcal{A}_{s_1} \| \leq T_{s1}
% \end{equation}
% then $x$ is assigned to the neuron $\mathcal{A}_{s_1}$, otherwise $x$ is wasted. If neuron $s_1$ and $s_2$ have edge, the age of the edge is set to zero,

\subsection{Anomaly Detection Rule}
This paper, assumes that the normal samples obey the Gaussian distribution and then leverages Mahalanobis distance to generate an anomaly score to each patch embedding. Defard \emph{et al.} \cite{defard2020padim} first propose this distance measure in anomaly segmentation, but they restrict each patch embedding to the corresponding position in the image. When input images are not aligned strictly, modeling patch feature embeddings in fixed positions fails.  In order to address this issue, we apply a global search in the NG network. Given a test image $I$ of size $W\times H $, we use $x_{ij}$ to denote the feature embedding of patch in position $(i,j)$. To calculate the anomaly score of position $(i,j)$, the nearest neuron in the network is searched firstly by:
\begin{equation}
    s_1=\underset{\mathcal{A}_s \in \mathcal{A}}{\operatorname{argmin}}\| x_{ij} - \mathcal{A}_s\|_2 .
\end{equation}
Then the Mahalanobis distance is computed as follows:
\begin{equation}
    D(x_{ij})=\sqrt{(x_{ij}-\mathcal{A}_{s_1})^T\Sigma_{\mathcal{A}_{s_1}}^{-1}(x_{ij}-\mathcal{A}_{s_1})}.
\end{equation}
Consequently, the matrix of the Mahalanobbis distance $D=(D(x_{ij}))_{1<i<W,1<j<H}$ that composes an anomaly score map can be calculated. Positions with high scores indicate that this area is probably anomalous. The image-level anomaly score is defined as the maximum value of the anomaly map $D$.

\section{Experiments}
\label{experiments}
\subsection{Dataset}
{\bf MVTec:} We evaluate the performance of our method on the MVTec AD dataset \cite{MVTec} which is collected for industrial anomaly detection and segmentation. The whole dataset is divided into 15 sub-datasets; each contains a nominal-only training dataset and a test set encompassing  both normal and anomalous samples. To assess the segmentation performance of algorithms, the dataset provides anomaly ground truth masks of various defect types, respectively.

{\bf BTAD:} We proceed to measure the performance on the BTAD dataset, another challenging industrial anomaly detection dataset. It contains RGB images of three different industrial products. Similar to MVTec AD, the training set includes only normal images, while both normal and abnormal images are present in the test set.
For each anomalous image, BTAD also provides a pixel-wise ground truth mask for evaluating the segmentation performance of the model.

In light of the limited number of test images in the original dataset compared to the training images, which poses challenges for assessing few-shot online learning performance, we have modified the dataset to align with the evaluation requirements. Specifically, we have retained only ten normal images in the training set, while the remaining normal images have been transferred to the test set for evaluation purposes.
Following the modification, the MVTec AD test dataset comprises 4737 normal images and 1258 abnormal images. Similarly, the BTAD test dataset consists of 2220 normal images and 290 abnormal images, resulting in an average ratio of normal to abnormal data of 7.7:1.  Notably, the number of normal images in both datasets significantly surpasses that of abnormal images after modification. This imbalance is representative of real-life production lines, where the majority of industrial products are deemed normal or qualified, in contrast to those exhibiting anomalies.

\subsection{Evaluation Protocols}
Before the evaluation process, the entire test set is randomly shuffled. Subsequently, images are sequentially selected from the shuffled test set to construct one online session after another. Each online session comprises an equal number of test images, and evaluation metrics are computed after collecting all the predictions within one session from the model. Following these operations, we can ensure that the test images have not been used for training prior to detection when evaluating the performance of our models on the online data stream. Meanwhile, the randomly mixing of normal and abnormal images in the data flow session avoids the centralization of defects and simulates the real ratio in the production line, where normal data significantly outweighs abnormal data. 
The final evaluation score is determined by computing the average detection accuracy across all sessions, serving as a metric to assess the overall performance of the model in handling the continuous flow of data. To facilitate a fair comparison, the random shuffle seeds are consistent across different methods.

For comprehensively assessing the few-shot learning and online learning capacity of models for dealing with data streams, the evaluation entails both offline and online settings.

{\bf Online evaluation:}
Models are trained using a few-shot training dataset and subsequently leverage the online data stream to autonomously update their parameters based on the detection results.

{\bf Offline evaluation:} 
In the offline setting, the approaches are exclusively trained on the few-shot training dataset and do not utilize data flow for parameter updates. This setting is compared to online evaluation and can assess the effectiveness of the model in learning from online data. Additionally, this evaluation setting can also reflect the ability to learn from few-shot samples to some extent.

\subsection{Evaluation Metrics}
Image-level anomaly detection performance is evaluated via the area under the receiver operator curve (ROCAUC), applying the anomaly scores computed by our method. Pixel-level anomaly segmentation performance is measured by two metrics. The first is pixel-level ROCAUC which is computed by scanning over the threshold on the anomaly score maps. However, previous work has proposed that ROCAUC can be biased in favor of large anomaly areas. To address this issue, Bergmann \emph{et al.} \cite{MVTec} propose to employ the per-region overlap (PRO) curve metric. This metric takes into account each anomalous connected component respectively to better account for diverse anomaly sizes. In detail, they scan over false positive rates (FPR) from 0 to 0.3 and compute separately the proportion of pixels of each anomalous connected component that is detected. The final score is the normalized integral area under the average PRO variation curve.

\begin{figure*}[!t]
\centering
\begin{tabular}{cccccc}
	\begin{tikzpicture}[/pgfplots/width=0.5\linewidth,/pgfplots/height=0.41\linewidth]
	\begin{axis}[ymin=70,ymax=100,xmin=-0.2,xmax=10.2,
	font=\scriptsize,
	title={\small{(a) Capsule}},
	title style={yshift=-4pt},
	xlabel={Sessions},
	ylabel={Accuracy~[\%]},
	ylabel near ticks,
	ylabel shift={-2pt},
	yticklabel style={/pgf/number format/fixed,/pgf/number format/precision=2},
	xtick={0,1,2,3,4,5,6,7,8,9,10},
	legend style={legend columns=1,font=\tiny},
	legend cell align={left},
	legend pos=south east,
	grid=both,
	grid style=dotted,
	major grid style={white!20!black},
	minor grid style={white!70!black},
	axis equal image=false]
	\addplot[mygreen,mark=*,mark size=1.2pt] coordinates{(0, 79.8) (1, 85.8) (2, 82.5) (3, 89.3) (4, 87.1) (5, 87.3) (6, 88.0) (7, 88.1) (8, 89.7) (9, 88.5) (10, 90.7)};
	\addplot[myblue,mark=*,mark size=1.2pt]  coordinates{(0, 98.1) (1, 98.5) (2, 98.6) (3, 98.8) (4, 98.6) (5, 98.9) (6, 98.9) (7, 98.8) (8, 98.8) (9, 99.0) (10, 98.5)};
	\addplot[myred,mark=*,mark size=1.2pt]   coordinates{(0, 91.7) (1, 92.4) (2, 94.1) (3, 93.8) (4, 92.0) (5, 94.3) (6, 92.8) (7, 93.0) (8, 93.2) (9, 94.6) (10, 94.7)};
	\legend{ROCAUC-image, ROCAUC-pixel, PRO}
	\end{axis}
	\end{tikzpicture} &
	\begin{tikzpicture}[/pgfplots/width=0.5\linewidth,/pgfplots/height=0.41\linewidth]
	\begin{axis}[ymin=70,ymax=100,xmin=-0.5,xmax=12.5,
	font=\scriptsize,
	title={\small{(b) Carpet}},
	title style={yshift=-4pt},
	xlabel={Sessions},
	ylabel={Accuracy~[\%]},
	ylabel near ticks,
	ylabel shift={-2pt},
	yticklabel style={/pgf/number format/fixed,/pgf/number format/precision=2},
	xtick={0,1,2,3,4,5,6,7,8,9,10,11,12},
	legend style={legend columns=1,font=\tiny},
	legend cell align={left},
	legend pos=south east,
	grid=both,
	grid style=dotted,
	major grid style={white!20!black},
	minor grid style={white!70!black},
	axis equal image=false]
	\addplot[mygreen,mark=*,mark size=1.2pt] coordinates{(0, 78.3) (1, 87.4) (2, 91.5) (3, 94.4) (4, 97.0) (5, 96.0) (6, 97.3) (7, 96.7) (8, 97.1) (9, 96.4) (10, 96.4) (11, 98.2) (12, 97.9)};
	\addplot[myblue,mark=*,mark size=1.2pt]  coordinates{(0, 96.2) (1, 98.3) (2, 98.7) (3, 99.0) (4, 99.0) (5, 99.0) (6, 99.1) (7, 99.2) (8, 99.1) (9, 99.2) (10, 99.2) (11, 99.1) (12, 99.4)};
	\addplot[myred,mark=*,mark size=1.2pt]   coordinates{(0, 87.8) (1, 93.6) (2, 95.1) (3, 96.1) (4, 96.4) (5, 96.3) (6, 96.9) (7, 96.6) (8, 96.6) (9, 96.3)(10, 96.6) (11, 96.9) (12, 96.8)};
	\legend{ROCAUC-image, ROCAUC-pixel, PRO}
	\end{axis}
	\end{tikzpicture} \\

	\begin{tikzpicture}[/pgfplots/width=0.5\linewidth,/pgfplots/height=0.41\linewidth]
	\begin{axis}[ymin=70,ymax=100,xmin=-0.2,xmax=4.2,
	font=\scriptsize,
	title={\small{(c) Toothbrush}},
	title style={yshift=-4pt},
	xlabel={Sessions},
	ylabel={Accuracy~[\%]},
	ylabel near ticks,
	ylabel shift={-2pt},
	yticklabel style={/pgf/number format/fixed,/pgf/number format/precision=2},
	xtick={0,1,2,3,4},
	legend style={legend columns=1,font=\tiny},
	legend cell align={left},
	legend pos=south east,
	grid=both,
	grid style=dotted,
	major grid style={white!20!black},
	minor grid style={white!70!black},
	axis equal image=false]
	\addplot[mygreen,mark=*,mark size=1.2pt] coordinates{(0,72) (1,88.5) (2,94.5) (3,96) (4,96.5)};
	\addplot[myblue,mark=*,mark size=1.2pt]  coordinates{(0,94) (1,97) (2,97.2) (3,97.5)(4,98)};
	\addplot[myred,mark=*,mark size=1.2pt]   coordinates{(0,74.8) (1,87) (2,91) (3,91.1) (4,90.9)};
	\legend{ROCAUC-image, ROCAUC-pixel, PRO}
	\end{axis}
	\end{tikzpicture} &

		\begin{tikzpicture}[/pgfplots/width=0.5\linewidth,/pgfplots/height=0.41\linewidth]
	\begin{axis}[ymin=30,ymax=100,xmin=-0.2,xmax=9.2,
	font=\scriptsize,
	title={\small{(d) Grid}},
	title style={yshift=-4pt},
	xlabel={Sessions},
	ylabel={Accuracy~[\%]},
	ylabel near ticks,
	ylabel shift={-2pt},
	yticklabel style={/pgf/number format/fixed,/pgf/number format/precision=2},
	xtick={0,1,2,3,4,5,6,7,8,9},
	legend style={legend columns=1,font=\tiny},
	legend cell align={left},
	legend pos=south east,
	grid=both,
	grid style=dotted,
	major grid style={white!20!black},
	minor grid style={white!70!black},
	axis equal image=false]
	\addplot[mygreen,mark=*,mark size=1.2pt] coordinates{(0, 51.5) (1, 53.6) (2, 57.9) (3, 55.4) (4, 61.7) (5, 58.1) (6, 60.8) (7, 58.7) (8, 57.6) (9, 66.3)};
	\addplot[myblue,mark=*,mark size=1.2pt]  coordinates{(0, 84.8) (1, 88.8) (2, 90.9) (3, 92.6) (4, 92.9) (5, 93.9) (6, 94.2) (7, 94.4) (8, 94.3) (9, 94.7)};
	\addplot[myred,mark=*,mark size=1.2pt]   coordinates{(0, 58.3) (1, 67.9) (2, 74.2) (3, 78.1) (4, 76.3) (5, 81.6) (6, 81.8) (7, 82.2) (8, 83.8) (9, 83.3)};
	\legend{ROCAUC-image, ROCAUC-pixel, PRO}
	\end{axis}
	\end{tikzpicture} \\

	\begin{tikzpicture}[/pgfplots/width=0.5\linewidth,/pgfplots/height=0.41\linewidth]
	\begin{axis}[ymin=50,ymax=90,xmin=-0.2,xmax=8.2,
	font=\scriptsize,
	title={\small{(e) Transistor}},
	title style={yshift=-4pt},
	xlabel={Sessions},
	ylabel={Accuracy~[\%]},
	ylabel near ticks,
	ylabel shift={-2pt},
	yticklabel style={/pgf/number format/fixed,/pgf/number format/precision=2},
	xtick={0,1,2,3,4,5,6,7,8},
	legend style={legend columns=1,font=\tiny},
	legend cell align={left},
	legend pos=south east,
	grid=both,
	grid style=dotted,
	major grid style={white!20!black},
	minor grid style={white!70!black},
	axis equal image=false]
	\addplot[mygreen,mark=*,mark size=1.2pt] coordinates{(0, 69.8) (1, 75.1) (2, 76.0) (3, 79.7) (4, 78.6) (5, 80.6) (6, 77.5) (7, 78.3) (8, 79.9)};
	\addplot[myblue,mark=*,mark size=1.2pt]  coordinates{(0, 84.8) (1, 81.6) (2, 86.0) (3, 84.3) (4, 84.4) (5, 86.7) (6, 85.7) (7, 85.8) (8, 87.4)};
	\addplot[myred,mark=*,mark size=1.2pt]   coordinates{(0, 67.5) (1, 64.6) (2, 71.0) (3, 65.7) (4, 68.2) (5, 71.0) (6, 71.7) (7, 70.8) (8, 73.7)};
	\legend{ROCAUC-image, ROCAUC-pixel, PRO}
	\end{axis}
	\end{tikzpicture} &

	\begin{tikzpicture}[/pgfplots/width=0.5\linewidth,/pgfplots/height=0.41\linewidth]
	\begin{axis}[ymin=70,ymax=100,xmin=-0.2,xmax=10.2,
	font=\scriptsize,
	title={\small{(f) Metal Nut}},
	title style={yshift=-4pt},
	xlabel={Sessions},
	ylabel={Accuracy~[\%]},
	ylabel near ticks,
	ylabel shift={-2pt},
	yticklabel style={/pgf/number format/fixed,/pgf/number format/precision=2},
	xtick={0,1,2,3,4,5,6,7,8,9,10},
	legend style={legend columns=1,font=\tiny},
	legend cell align={left},
	legend pos=south east,
	grid=both,
	grid style=dotted,
	major grid style={white!20!black},
	minor grid style={white!70!black},
	axis equal image=false]
	\addplot[mygreen,mark=*,mark size=1.2pt] coordinates{(0, 82.2) (1, 89.5) (2, 91.7) (3, 94.1) (4, 93.4) (5, 95.7) (6, 95.0) (7, 95.1) (8, 94.5) (9, 95.5) (10, 96.1)};
	\addplot[myblue,mark=*,mark size=1.2pt]  coordinates{(0, 89.6) (1, 88.7) (2, 90.1) (3, 89.9) (4, 89.1) (5, 89.6) (6, 89.1) (7, 89.3) (8, 89.5) (9, 89.3) (10, 90.0)};
	\addplot[myred,mark=*,mark size=1.2pt]   coordinates{(0, 78.2) (1, 79.9) (2, 83.7) (3, 82.8) (4, 81.3) (5, 85.1) (6, 83.1) (7, 82.8) (8, 85.3) (9, 84.7) (10, 84.7)};
	\legend{ROCAUC-image, ROCAUC-pixel, PRO}
	\end{axis}
	\end{tikzpicture} \\
	
\end{tabular}
\caption{The accuracy curve of  sub-datasets on MVTec AD dataset.  We report Image-level (ROCAUC \%),  Pixel-level (ROCAUC \%) and Per-region overlap (PRO \%) of each session.}
\label{trend}
\end{figure*}

\subsection{Comparison results}
We first present the comparative results of our methods on the MVTec dataset. MVTec is specifically designed for anomaly detection and segmentation, thus we report a complete set of results, including Image-level (ROCAUC \%),  Pixel-level (ROCAUC \%), and per-region overlap (PRO \%).  To investigate the online learning effects, the variation curve of detection accuracy is depicted in Figure \ref{trend}.

Due to the random selection of online sessions and the varying degrees of difficulty for different types of anomalies, the curve exhibits significant fluctuations. To address this, we adjust the batch size within each session and conduct 20 runs of our method for computing an average to achieve a smoother curve.  As shown in Figure \ref{trend}, the accuracy of anomaly detection and segmentation shows a continuous upward trend overall. The image-level ROCAUC and per-region overlap exhibit relatively dramatic increases; for example, the image-level ROCAUC and PRO for "Carpet" increase by approximately 20\% and 10\% respectively throughout the entire process. In contrast, the pixel-level ROCAUC curve appears relatively smooth. This phenomenon is attributed to a significant imbalance in the number of normal pixels and abnormal pixels, making it challenging to improve the performance on this metric.

We then proceed to compare our method with other prevalent approaches in anomaly detection and segmentation. As shown in Table \ref{newone}, our method demonstrates a significant improvement over previous work, leveraging online learning to achieve an image-level ROCAUC of 87.4 and a PRO score of 88.1. In contrast, approaches lacking online learning capabilities only achieve an image-level ROCAUC of 85.4 and a PRO score of 86.7, as demonstrated by the PatchCore method which utilizes a more intricate neighborhood aggregation feature representation; This outcome underscores the robust online learning capabilities of our approach. Detailed results for each category in the MVTec dataset are presented in Table \ref{detail_newone}.

\begin{table*}[!ht]\scriptsize
\centering
 \caption{Comparison results on MVTec dataset under few-shot online detection setting, we report average results of all sessions: Image (Image-level (ROCAUC \%)), PRO (per-region overlap (PRO \%))}
\renewcommand{\arraystretch}{1.2}
\tabcolsep=1pt
 \begin{tabular}{l|ccccccc}
 \toprule
 \multicolumn{7}{c}{Offline Methods}\\
   \midrule
  \bfseries Method & DRAEM~\cite{zavrtanik2021draem} &  DiffusionAD~\cite{zhang2023diffusionad} &RegAD-L~\cite{huang2022registration}& SPADE~\cite{spade} & PaDim~\cite{defard2020padim} &  PatchCore ~\cite{roth2021towards} & Ours \\

\midrule
 \bfseries Image & 82.2 & 84.4  & 84.9 & 77.4 & 77.4 &85.4&78.2\\
  \bfseries PRO & 76.6 & 78.8 & 85.2& 83.1 & 82.6 &86.7&82.8\\
  \midrule
  \multicolumn{7}{c}{Online Methods}\\
  \midrule
   \bfseries Image & 86.2 & 86.8 & 80.1 &80.3 &77.7 & 84.9 &\bf 87.4\\
  \bfseries PRO & 78.9 & 80.0 & 81.3  &85.1 & 83.1 & 85.8 & \bf 88.1\\
 \bottomrule
 \end{tabular}
   \label{newone}
\end{table*}

Furthermore, to delve deeper into the performance of online learning, we conducted an evaluation of our method under the offline learning procedure, with the comparison results presented in Table \ref{newone}. The offline model yields 78.2 and 82.8 in image-level ROCAUC and PRO metrics, though the result is not the best within offline methods, just falling between PaDiM and SPADE, it can achieve a much higher performance after being reinforced by the online learning process. 

We then modify PaDiM, SPADE, DRAEM,  DiffusionAD and PatchCore to incorporate the capability of online learning: wherein the feature embedding/image is added to the model for parameter updates if the image anomaly score falls below a fixed threshold. PaDiM is almost the same as its offline version, this is because that  PaDim divides an image into a grid of $(i,j)\in[1,W]\times[1,H]$ and associates a multivariate Gaussian distribution to fixed position $(i,j)$. However, images in the MVTec AD dataset are not strictly aligned, which makes the distribution of position $(i,j)$ contains appearance information from different parts of the object. This inherent bias results in the model having difficulty adjusting the distribution parameters accurately through online learning. 
The online version of SPADE exhibits only a marginal performance improvement. It is important to note that while SPADE shows some progress, the method requires constant enlargement of its memory bank, potentially exceeding device storage limits. This point also proves the effectiveness of our method in another way. DRAEM demonstrates some performance improvement compared to other methods in the comparison, but it still lags significantly behind our approach.  DRAEM is a deep neural network model that tends to overfit current data and lose previously learned information \cite{wei2023topology}, which makes it hard to learn a comprehensive image feature distribution. Benefiting from the sophisticated local-aware patch features that aggregate information from the neighborhood, PatchCore performs well in offline settings with just a few training samples. However, it also does not improve detection ability through the online learning stage. A possible reason for this phenomenon is that this algorithm is easy to sample misjudged outliers into the coreset during online learning. Outliers typically reside at a considerable distance from the normal coreset in the feature space, and are more likely to be incorporated into the coreset due to the minimax facility location coreset selection algorithm which selects the farthest point from the current coreset. As a result, the presence of outliers in the coreset notably impacts the learning process and the performance of memory bank-based algorithms.

\begin{figure}[!ht]
    \centering
    \includegraphics[width=0.45\textwidth]{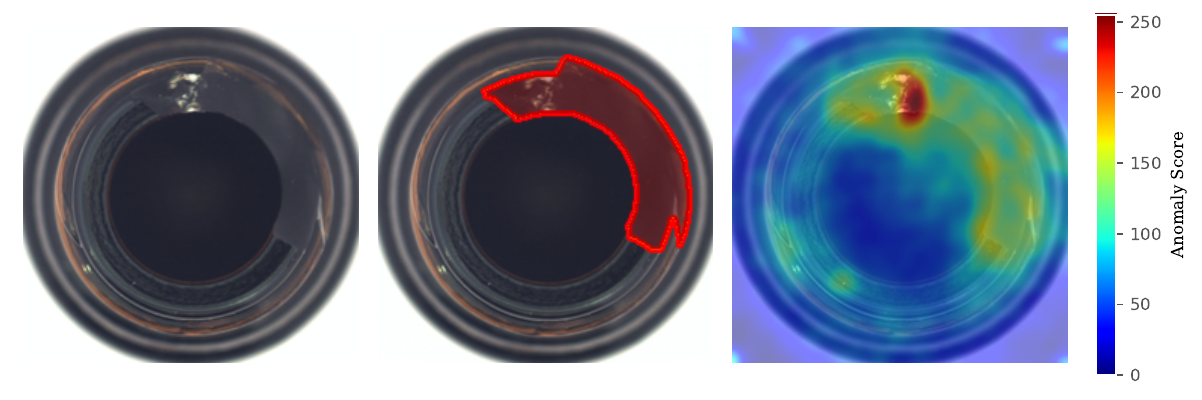}
    \includegraphics[width=0.45\textwidth]{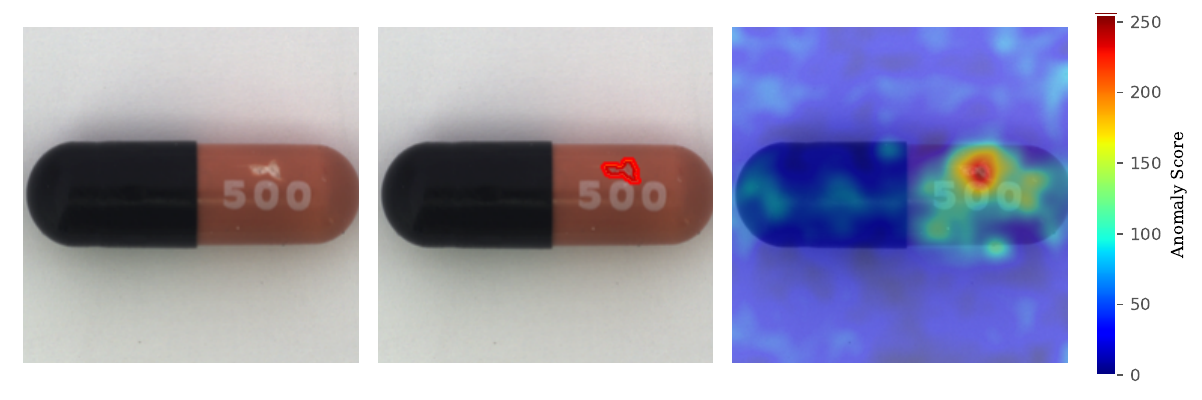}
    \includegraphics[width=0.45\textwidth]{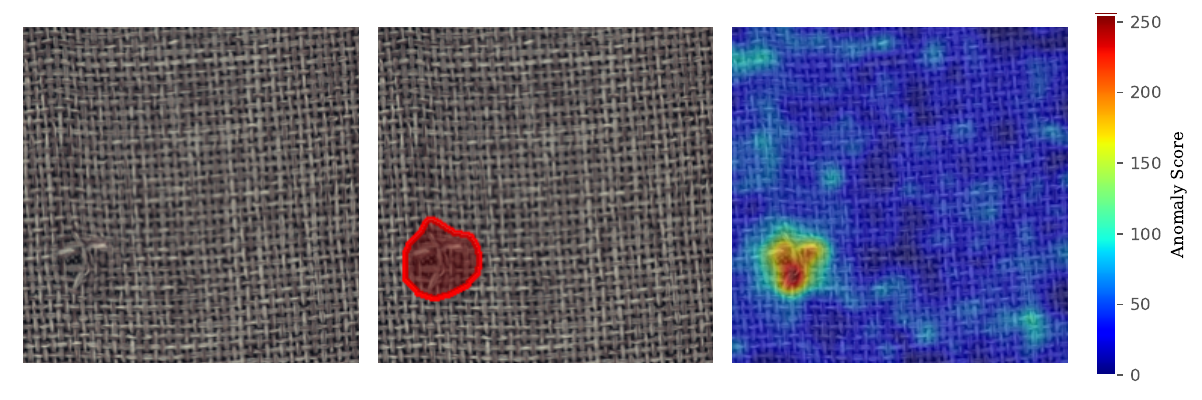}
     \includegraphics[width=0.45\textwidth]{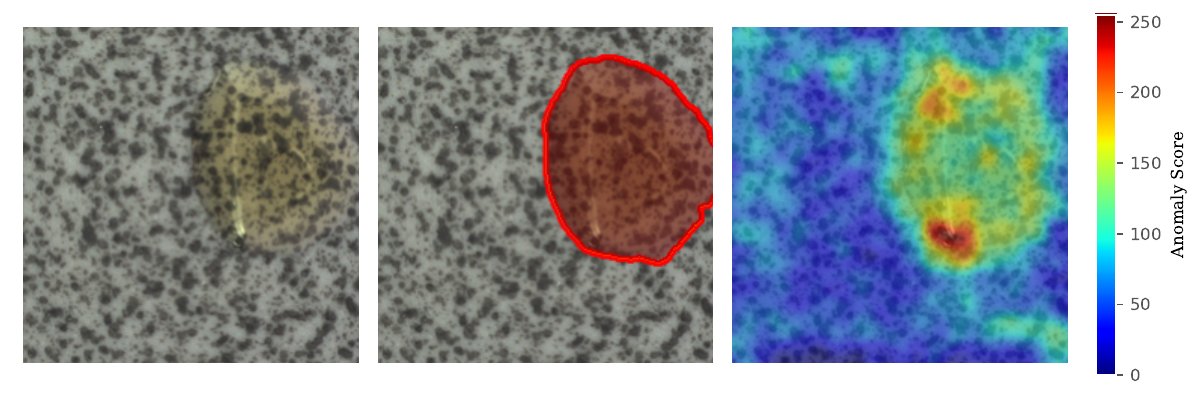}
    \includegraphics[width=0.45\textwidth]{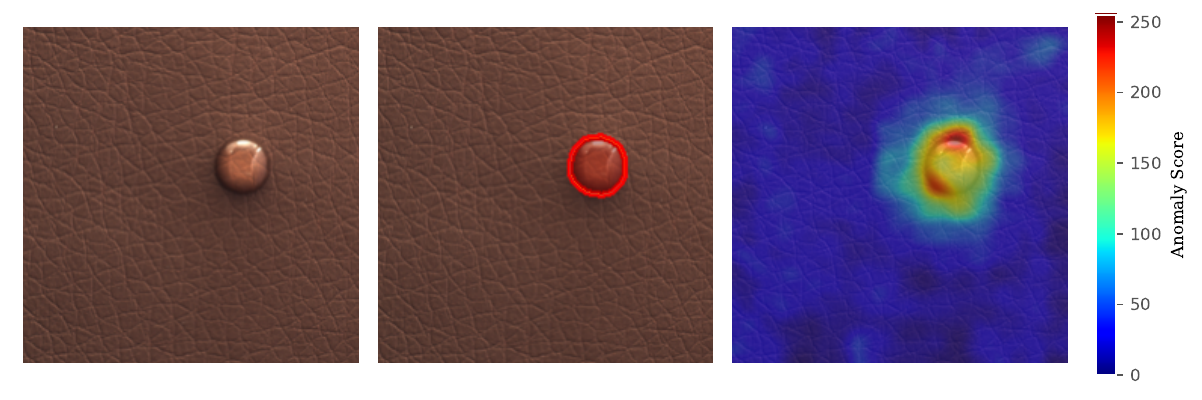}
    \includegraphics[width=0.45\textwidth]{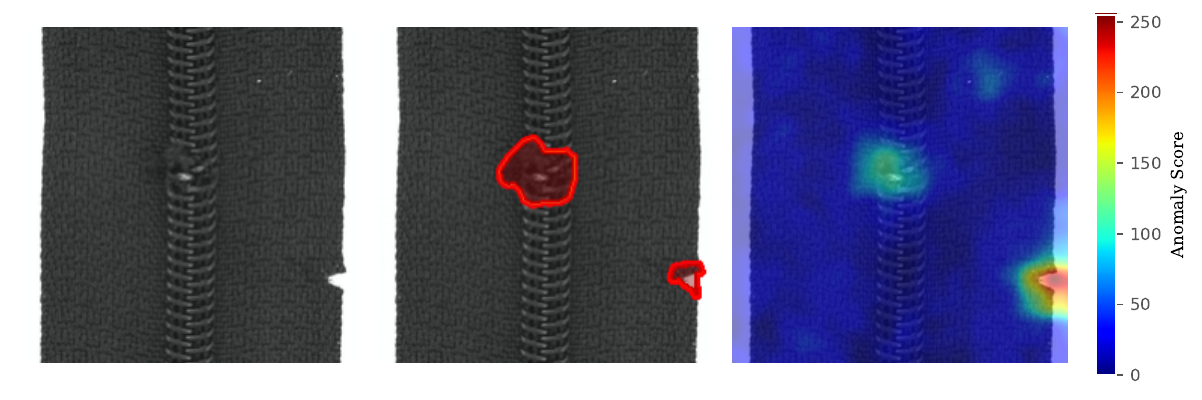}
    
    \caption{Qualitative results of our anomaly segmentation method on
the MVTec dataset. {\bf Left column:} Input Images containing anomaly regions. {\bf Center column:} Ground truth regions of anomalies. {\bf Right column:} Anomaly Score maps for each test image predicted by our algorithm.}
    \label{vis}
\end{figure}

\begin{figure}[!ht]
    \centering
    \includegraphics[width=1.0\textwidth]{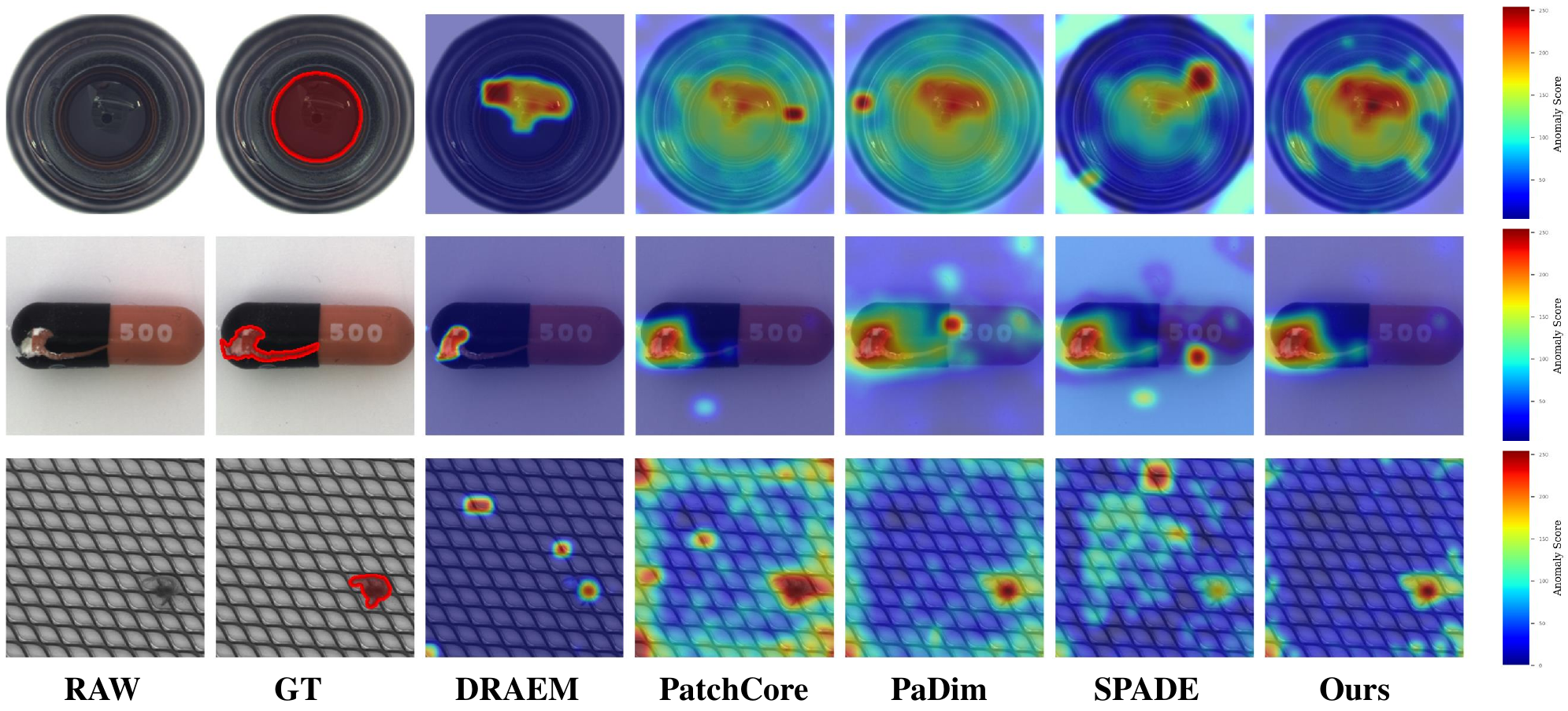}
    
    \caption{\textbf{Qualitative results.} We further visualize the results of the proposed and other conventional approaches. Raw indicates the original images and GT means the visualization of ground truth segmentation.}
    \label{vis_comp}
\end{figure}

Finally, we provide examples of anomaly score maps generated by our and other conventional methods on the MVTec AD dataset in Figure \ref{vis}, \ref{vis_comp}. These heat maps show the segmentation results of the last network trained by the data flow.

The performance of our algorithm on the BTAD dataset is reported in Figure \ref{BTADfig}, with the evaluation protocol mirroring the experiments conducted on the MVTec dataset. We report the image-level (ROCAUC \%)  and per-region overlap anomaly detection results. The bar chart demonstrates that our method achieves an ROCAUC of 86.5 and a PRO score of 92.8, securing the top position in both image detection and pixel segmentation. Although the SPADE and PaDim obtain a near precision in PRO and image-level ROCAUC, respectively, they both fall short in the other task.

\begin{figure}[!ht]
\centering
\begin{tabular}{l}
\begin{tikzpicture}[/pgfplots/width=0.7\linewidth,/pgfplots/height=0.5\linewidth]
\begin{axis}
[
    xbar, % ybar command displays the graph in horizontal form, while the xbar command displays the graph in vertical form.
    bar width=10pt,
    enlarge y limits=0.5,
    xmin=75,
    xmax=95,
    % enlargelimits=0.15, % these limits are used to shrink or expand the graph. The lesser the limit, the higher the graph will expand or grow. The greater the limit, the more graph will shrink. 
    legend style={at={(0.5,-0.25)},font=\scriptsize,
    legend cell align={center},% these are the measures of the bottom row containing surplus (wheat, Tea, rice), where -0.25 is the gap between the bottom row and the graph. 
      anchor=north, legend columns=-1}, % here, north is the position of the bottom legend row. You can specify the east, west, or south direction to shift the location. 
    xlabel={Accuracy~[\%]}, % there should be no line gap between the rows here. Otherwise, latex will show an error.
    symbolic y coords={Image, PRO}, 
    ytick=data, 
    % ytick distance=0.1，
    nodes near coords, 
    nodes near coords style={/pgf/number format/.cd, fixed zerofill,precision=1},
    nodes near coords align={horizontal},font=\scriptsize ]
\addplot[color=orange,fill=orange] coordinates {(92.8,Image) (82.2,PRO)};
\addplot[color=myblue,fill=,myblue] coordinates {(88.9,Image) (84.7,PRO)}; 
\addplot[color=mygreen,fill=mygreen] coordinates {(91.1,Image) (76.0,PRO)};
\addplot[color=mypurple,fill=mypurple] coordinates {(93.0,Image) (86.5,PRO)};% these are the measures of a particular bar graph. The tick marks of the y-axis will be adjusted automatically according to the data values entered in the coordinates.

\legend {PaDim,SPADE,PatchCore,Ours}

\end{axis}
\end{tikzpicture}\\
	\end{tabular}
\caption{Comparison of our method with other offline methods on BTAD dataset. We report Image (Image-level (ROCAUC \%)) and PRO (Per-region overlap (PRO \%)) }
\label{BTADfig}
\end{figure}

\begin{table}[!hb]
\centering
  \caption{Mean inference time per image and GPU memory footprint of our method on MVTec AD. Scores are (Image ROCAUC, PRO metric).}
 \renewcommand{\arraystretch}{1}
 \centering
 \begin{tabular}{lcc}
 \toprule
 \bfseries Model & \bfseries Online  &\bfseries Offline  \\
 \midrule
 \bf Time (sec.) & 0.839    & 0.514\\
 \bf GPU Memory (MB) & 2217    &  2217 \\
 \bf Scores & (87.4, 88.1)  & (78.2, 82.8) \\
 \bottomrule
 \end{tabular}
  \label{inference}
\end{table}
\subsection{Inference Time}
Inference time is another dimension which we are interested in. We report the results in  Table \ref{inference}. We measure the model inference time on the CPU (Intel i7-4770 @3.4GHz) with a GPU (NVIDIA TITAN X (Pascal)). All methods leverage GPU operations to accelerate the algorithm whenever possible. The statistics, including the online learning stage, are displayed in Table \ref{inference}. Although our method updates its parameters to enhance accuracy, it just consumes 0.315 seconds more than the algorithm without online learning ability. Generally, our model can process an image within one second while continuously updating itself. {Besieds, shown in the Table, our method only consumes around 2.2GB GPU memory, which can run on most consumer-grade GPU cards.}

\begin{figure}[t]
\centering
\begin{tabular}{l}
	\begin{tikzpicture}[/pgfplots/width=0.8\linewidth,/pgfplots/height=0.5\linewidth]
	\begin{axis}[ymin=70,ymax=100,xmin=log10(0.7),xmax=log10(40),
	font=\scriptsize,
	title={},
	title style={yshift=-4pt},
	xlabel={$log10(N)$},
	ylabel={Accuracy~[\%]},
	ylabel near ticks,
	ylabel shift={-2pt},
	yticklabel style={/pgf/number format/fixed,/pgf/number format/precision=2},
	xtick={log10(0.7),log10(1),log10(2),log10(5),log10(10),log10(20),log10(30),log10(40)},
	xticklabels={lg0,lg1,lg2,lg5,lg10,lg20,lg30,lg40},
	%ytick={0,0.02,...,1},
% 	 symbolic x coords={lg0,lg1,lg2,lg5,lg10,lg20,lg30,lg40}, xtick={log10(0.1),log10(1),log10(2),log10(5),log10(10),log10(20),log10(40)},
	legend style={legend columns=1,font=\tiny},
	legend cell align={left},
	legend pos= south east,
	grid=both,
	grid style=dotted,
	major grid style={white!20!black},
	minor grid style={white!70!black},
	axis equal image=false]
	\addplot[mygreen,mark=*,mark size=1.2pt] coordinates{ (log10(0.7),72.9) (log10(1),86.4) (log10(2),87.2) (log10(5),87.2) (log10(10),87.4) (log10(20),86.9) (log10(30),86.9) (log10(40),86.9)};
	\addplot[myblue,mark=*,mark size=1.2pt]  coordinates{ (log10(0.7),94.1)(log10(1),95.4) (log10(2),95.4) (log10(5),95.4) (log10(10),95.4) (log10(20),95.4) (log10(30),95.4) (log10(40),95.4)};
	\addplot[myred,mark=*,mark size=1.2pt]   coordinates{(log10(0.7),85.3) (log10(1),87.7) (log10(2),87.9) (log10(5),87.9) (log10(10),88.1) (log10(20),88.0) (log10(30),88.0) (log10(40),88.0)};
	\legend{ROCAUC-image, ROCAUC-pixel, PRO}
	\end{axis}
	\end{tikzpicture} \\
	\end{tabular}
\caption{Ablation study of the epoch $N$  on MVTec AD. We report Image-level (ROCAUC \%),  Pixel-level (ROCAUC \%) and Per-region overlap (PRO \%) }
\label{Nablation}
\end{figure}

\subsection{Ablation Study}
In this section, we first investigate the performance under different hyper-parameter settings. The network comprises three hyper-parameters: the initial cluster epoch $N$, the number of the cluster centers $K$, and the maximum age $age_{max}$ of the edge in the network. The hyper-parameters used for comparison with other methods are highlighted with gray background in the table.

The results of hyper-parameter cluster epoch $N$ are reported in Figure \ref{Nablation}; we fixed other hyper-parameters and altered the number of the epochs.  It is evident that increasing the epoch number $N$ has limited effects on the final result. The per-pixel ROCAUC and per-region overlap are almost identical, while $N$ varies from 10 to 40. The variation range of image-level ROCAUC is slightly more extensive.  Notably, even with only one epoch, the performance is quite high, indicating that the initial algorithm has a relatively fast convergence rate.

\begin{table}[hb]
\small
\centering
 \caption{Ablation study of the initial cluster size $K$  on MVTec. We report Image (Image-level (ROCAUC \%)),  Pixel (Pixel-level (ROCAUC \%)) and PRO (per-region overlap (PRO \%))}
 \renewcommand{\arraystretch}{1}
 \begin{tabular}{lccc>{\columncolor{gray!20}}cc}
 \toprule
 \bfseries $K$ &14$\times$ 14&  28$\times$ 28   & 42$\times$ 42 &56$\times$ 56 &70$\times$ 70\\
 \midrule
 \bfseries Image  &82.6   &\bf 87.4    &86.8   &\bf 87.4  & 87.1\\
 \bfseries Pixel  &93.7   &95.2  & 95.3   & 95.4 & \bf 95.5\\
 \bfseries PRO & 82.9     &86.9  & 87.9  & 88.1 & \bf88.2 \\
 \bottomrule
 \end{tabular}
   \label{Kablation}
\end{table}

We then proceed with ablation studies on the size of the initial cluster $K$. The experiment results are presented in Table \ref{Kablation}. Generally, the segmentation performance improves with the enlargement of the initial cluster size. The rate of improvement is particularly rapid when the size is small and becomes slower after the size reaches $42 \times 42$. Nevertheless, the trend of the image-level ROCAUC is unusual. It reaches an extreme value at $28 \times 28$ size and then declines until reaching another one. We will discuss this phenomenon later in Section \ref{dis}. To strike a balance between accuracy and computational complexity, $K= 56\times 56$ is chosen in this paper.

We also research the significance of the maximum age $Age_{max}$ of the edge in the network. Epoch $N$ and the number of the cluster center $K$ are fixed during the studies. The results, shown in Table \ref{Ageablation}, include Image-level ROCAUC, Pixel-level ROCAUC, and Per-region overlap under a series of $Age_{max}$. Similar to the number of epoch $N$, the Pixel-level ROCAUC and the Per-region overlap are almost the same no matter how  $Age_{max}$ changes. Only image-level ROCAUC varies from 86.8 to 87.4 in the research, and after the parameter exceeds 100, the accuracy stabilizes.
\begin{table}[!ht]\small
\centering
 \caption{Ablation study of the maximum age $age_{max}$ of the edge  on MVTec. We report Image (Image-level (ROCAUC \%)),  Pixel (Pixel-level (ROCAUC \%)) and PRO (per-region overlap (PRO \%))}
 \renewcommand{\arraystretch}{1}
 \begin{tabular}{lcc>{\columncolor{gray!20}}ccccc}
 \toprule
 \bfseries $Age_{max}$ &5&10&25&50& 100 &  500  &  1500 \\
 \midrule
 \bfseries Image  &86.6&87.1& \bf 87.4&87.2& 86.9 & 86.7  & 86.7  \\
 \bfseries Pixel  &95.3& 95.3&\bf 95.4&\bf 95.4&\bf 95.4& \bf 95.4  &\bf 95.4\\
 \bfseries PRO & 87.7&87.9&\bf 88.1&88.1& 88.0 & 88.0  & 87.9 \\
 \bottomrule
 \end{tabular}
   \label{Ageablation}
\end{table}

{We also conduct ablation studies to investigate the effectiveness of the different types of thresholds. In this paper, we apply four types of thresholds: the mean distance between the neural node and its neighbors, the median distance to the neighbors, the maximum distance from the neural node to its neighbors, and the minimum distance. The results are presented in Table \ref{Tablation}, obviously, employing the median distance between the neural node and its neighbors yields the best performance in both detection and segmentation, while the mean distance achieves the second-highest performance in image-level detection. Unexpectedly, the model without any threshold gets 86.3 in image-level ROCAUC and outperforms the threshold using the maximum distance, which is proposed in video anomaly detection by previous work \cite{sun2017online}. These results demonstrate the superiority of our threshold system.}

\begin{table}[!ht]\small
\centering
 \caption{Ablation study of different types of threshold on MVTec AD. We report Image (Image-level (ROCAUC \%)),  Pixel (Pixel-level (ROCAUC \%)) and PRO (per-region overlap (PRO \%))}
 \renewcommand{\arraystretch}{1}
 \begin{tabular}{lc >{\columncolor{gray!20}}cccc}
 \toprule
  \bfseries Threshold Type &\bfseries Min &  \bfseries Median & \bfseries Mean& \bfseries Max & \bfseries None  \\
 \midrule
 \bfseries Image  &87.2 & \bf 87.4 &  87.3  & 86.1  & 86.3  \\
 \bfseries Pixel &94.7 &\bf  95.4 & 95.1  & 95.0  &  95.0 \\
 \bfseries PRO &87.1          &\bf 88.1 & 87.8   & 87.6  & 87.5  \\
 \bottomrule
 \end{tabular}
   \label{Tablation}
\end{table}

We also research the effect of the number of training samples. The results are reported in Table \ref{Sizeablation}. Despite our method using only half of the training dataset, it still reaches a comparable accuracy. The image-level ROCAUC and PRO still outperform the online version SPADE~\cite{spade} and PaDim~\cite{defard2020padim}. This result demonstrates the effect of our method on few-shot task again. 

{Additionally, we give parameter volume information in Table \ref{comp}. As shown in the table, our method maintains a good balance between performance improvement and parameter quantity. Although SPADE~\cite{spade} and PatchCore~\cite{roth2021towards} have smaller initial parameter sizes, they need to expand memory bank size constantly during the online learning stage, which poses a risk of storage overflow.}

\begin{table}[!ht]\small
\centering
 \caption{Ablation study of the size of training dataset on MVTec. We report Image-level (ROCAUC \%), Pixel-level (ROCAUC \%) and PRO (per-region overlap (PRO \%))}
 \renewcommand{\arraystretch}{1}
 \begin{tabular}{lccc}
 \toprule
  \bfseries Size &\bfseries Image-level &  \bfseries Pixel-level & \bfseries PRO \\
 \midrule
 \bfseries 5-shot & 85.3 & 94.9   & 87.2  \\
 \bfseries 10-shot &87.4 & 95.4  & 88.1  \\
 \bfseries 20-shot &88.1 & 95.8  & 89.4  \\
 \bfseries 30-shot &88.3 & 96.1  & 90.0  \\
 \bfseries 50-shot &89.8 & 96.4  & 90.4  \\
 \bottomrule
 \end{tabular}
   \label{Sizeablation}
\end{table}

\begin{table}[!ht]\small
\centering
 \caption{Approximate initial model parameters of the evaluation methods. ``Extra Memory" means the model needs to expand memory requirements continuously with the online learning process. ``Improvement" is (Image ROCAUC, PRO metric), which is the performance improvements provided by the online learning stage. }
 \renewcommand{\arraystretch}{1}
  \setlength{\tabcolsep}{3pt}
 \begin{tabular}{lccc}
 \toprule
  \bfseries Method &\bfseries Initial Parameters(M) &\bfseries Extra Memory&  \bfseries Improvement \\
 \midrule
 \bfseries DRAEM \cite{zavrtanik2021draem} &       93.1 & & \textcolor{green!70!black}{{(+4.0, +2.3)\%}}\\
 \bfseries DiffusionAD \cite{zhang2023diffusionad} &       152.5 & & \textcolor{green!70!black}{{(+2.4, +1.2)\%}}\\
 \bfseries RegAD-L\cite{huang2022registration} &    38.1 & \checkmark & \textcolor{red!70!black}{{(-4.8, -3.9)\%}}\\
 \bfseries PatchCore\cite{roth2021towards} &    11.7 & \checkmark & \textcolor{red!70!black}{{(-0.5, -0.9)\%}}\\
 \bfseries SPADE \cite{spade}&         12.5 & \checkmark& \textcolor{green!70!black}{{(+2.9, +2.0)\%}}\\
 \bfseries PaDim \cite{defard2020padim}&          41.9 & & \textcolor{green!70!black}{{(+0.3, +0.5)\%}}      \\
 \bfseries Ours &                 42.1  & &\textcolor{green!70!black}{{(+9.2, +5.3)\%}}   \\
 
 \bottomrule
 \end{tabular}
   \label{comp}
\end{table}

\subsection{Implementation Details}
In our experiments, ResNet18 which is pre-trained on the ImageNet dataset is employed as the feature extractor. All the images are resized to $256 \times 256$ and then cropped to $224 \times 224$. In order to achieve the multi-scale representation, we use features at the end of the first block, the second block, and the third block from the ResNet18 backbone. Feature maps of different sizes are resized to $224 \times 224$ by bilinear interpolation. The features from different blocks with the same position are concatenated together and reduced dimension to 100 by random selection. We selected 3136 cluster centers, and the training epoch is set to 10 in the initialization stage of the model. In the online learning stage, the batch size is set to 10. The hyper-parameter $age_{max}$ is set at 25.  After obtaining the pixel-level anomaly score, a Gaussian filter ($\sigma=4$) is employed to smooth the map. We reimplemented SPADE, PaDiM, DRAEM, DiffusionAD and PatchCore under FOADS setting to demonstrate our online performance. ResNet18 is used as the backbone network and the same feature dimension is applied in SPADE, PaDim, RegAD and PatchCore for a fair comparison. And environment with CUDA 10.2, Ubuntu 16.04, Python 3.7 is employed for the experiments in this paper.

\section{Discussion and Outlook}
\label{dis}
\subsection{Failure Cases and Limitations}
\begin{figure}[!ht]
    \centering
    \includegraphics[width=0.45\textwidth]{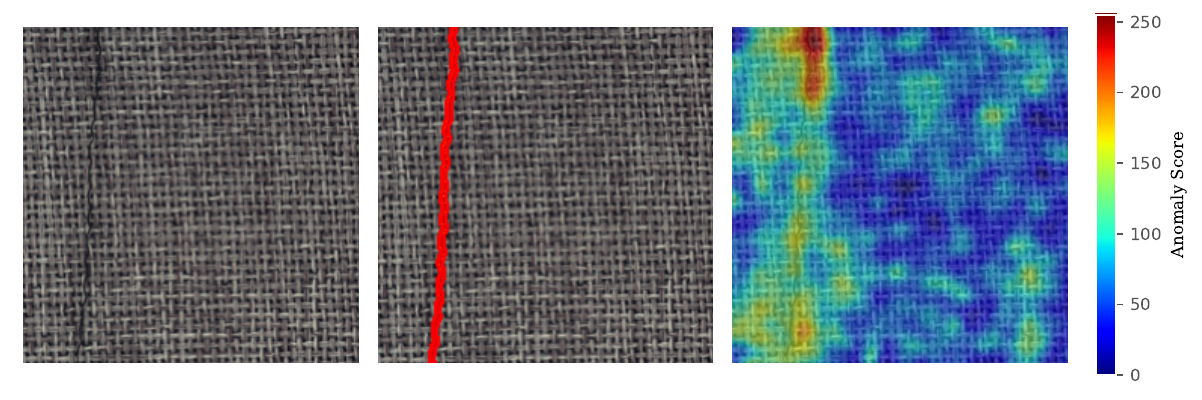}
    \includegraphics[width=0.45\textwidth]{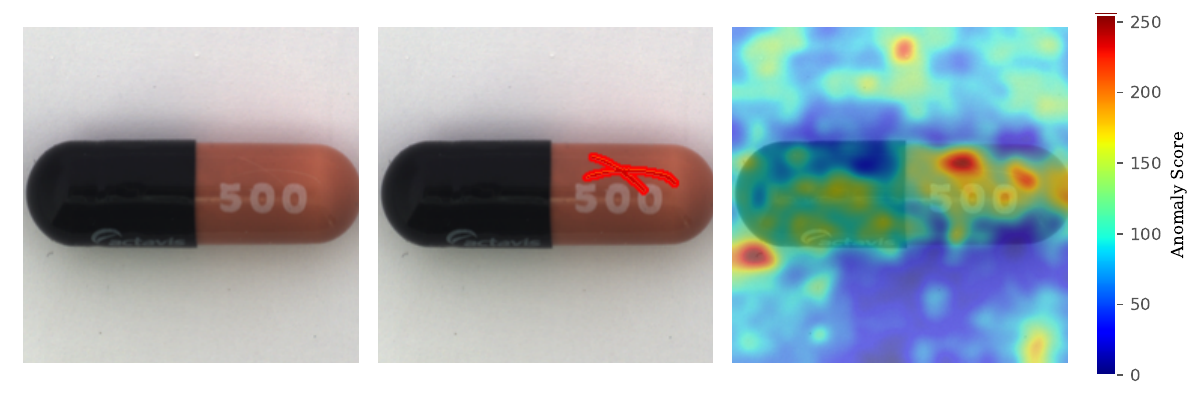}
    
    \caption{Visualization of failure cases. {\bf Left column:} Input Images containing anomaly regions. {\bf Center column:} Ground truth regions of anomalies. {\bf Right column:} Anomaly Score maps for each test image predicted by our algorithm.}
    \label{failure_case}
\end{figure}

{Shown in Figure \ref{failure_case}, when evaluated on the MVTec AD dataset, our model does not perform well on narrow and elongated type defects.This is mainly due to the presence of downsampling  in convolutional neural networks. The embeddings in the feature map represent a patch rather than a pixel. Thus, the segmentation results are not accurate enough to detect very minor anomalies.}

{Our method also has some limitations on the selection of initial cluster size. As mentioned in the ablation study, although there is a certain change trend when the hyper-parameter $K$ varies, the image-level ROCAUC differs. The main reason for this phenomenon is that some sub-classes in the MVTec AD exhibit an opposite trend with changeable hyper-parameter. For example, as shown in Figure \ref{dis_pic}, in the ablation study of the cluster size $K$, the ``Grid" sub-class shows a  significant difference compared to others. The peak of ``Grid" appears at $28 \times 28$ and then declines rapidly,  while others are still increasing. This may be attributed to ``Grid"  just consisting of simple texture patterns, which does not need too many clusters to represent. }

\subsection{Outlook}
In the future, it would be beneficial to make the establishment of the model more flexible, for example, the initial size can be self-tuned to suit different sub-classes. We can begin with only two nodes and determine whether new data belongs to existing clusters. If not, the model should grow the scale of the network for better fitting. This iterative growth process enables the model to adapt to the specific requirements of different sub-classes, resulting in an arbitrary-sized network that is tailored to address their respective peculiarities.

\begin{figure}[!ht]
\centering
\begin{tabular}{l}
	\begin{tikzpicture}[/pgfplots/width=0.6\linewidth,/pgfplots/height=0.5\linewidth]
	\begin{axis}[ymin=60,ymax=95,xmin=10,xmax=60,
	font=\scriptsize,
	title={},
	title style={yshift=-4pt},
	xlabel={$K$},
	ylabel={Accuracy~[\%]},
	ylabel near ticks,
	ylabel shift={-2pt},
	yticklabel style={/pgf/number format/fixed,/pgf/number format/precision=2},
	xtick={0,14,28,42,56},
	legend style={legend columns=1,font=\tiny},
	legend cell align={left},
	legend pos= outer north east,
	grid=both,
	grid style=dotted,
	major grid style={white!20!black},
	minor grid style={white!70!black},
	axis equal image=false]
	\addplot[mygreen,mark=*,mark size=1.2pt] coordinates{ (14,61) (28,70) (42,63) (56,60)};
	\addplot[myblue,mark=*,mark size=1.2pt]  coordinates{ (14,85) (28,92) (42,93) (56,94)};
	\addplot[myred,mark=*,mark size=1.2pt]   coordinates{(14,78) (28,82) (42,90) (56,92)};
	\addplot[orange,mark=*,mark size=1.2pt]   coordinates{(14,70) (28,81) (42,83) (56,81)};
	\legend{Grid, Metal, Zipper,Pill}
	\end{axis}
	\end{tikzpicture} \\
	\end{tabular}
\caption{The accuracy curve of some sub-datasets on MVTec AD dataset.  We report Image-level (ROCAUC \%) under different $K$.}
\label{dis_pic}
\end{figure}

\section{Conclusion}
\label{conclusions}
This paper, focuses on an unsolved, challenging, yet practical few-shot online anomaly detection and segmentation scenario. The task involves models updating themselves from unlabeled streaming data containing both normal and abnormal images, with only a few samples provided for initial training. We propose a framework based on Neural Gas (NG) network to model the multi-scale feature embedding extracted from CNN. Our NG network maintains the topological structure of the normal feature manifold to filter the abnormal data. Extensive experiments demonstrate that our method can enhance the accuracy as the unlabeled data flow constantly enters the system. Further study reveals that the time complexity of our model falls within an acceptable range.

\section{Appendices}
% \usepackage{multirow}
\begin{sidewaystable}[htbp]
\centering
\caption{Comparison results on MVTec dataset with ResNet18 under few shot online detection setting, we report average results of all sessions: Image (Image-level (ROCAUC \%)), PRO (per-region overlap (PRO \%))}
%\scriptsize
\renewcommand{\arraystretch}{1.2}
\setlength{\tabcolsep}{4pt}
\resizebox{\linewidth}{!}{
\begin{tabular}{rr|cccccccccccccccc}
%\hline
%                                        & \multicolumn{1}{l}{}                                                                        &        & \multicolumn{15}{c}{Category}                                                                                                             &         \\ \cline{2-19} 
\toprule
\multicolumn{2}{c|}{ Category} & {Bottle} & {  Cable} & { Capsule} & { Carpet} & { Grid} & { Hazelnut} & { Leather} & { Metal Nut} & { Pill} & { Screw} & { Tile} & { Toothbrush} & { Transistor} & { Wood} & { Zipper} & { Average} \\
\midrule
\multicolumn{18}{c}{Offline Methods}\\
\midrule
\multicolumn{1}{r|}{\multirow{3}{*}{{DRAEM~\cite{zavrtanik2021draem}}}}  & Image &  99.6 &  76.7  &  69.2 & 88.9  &  65.4    &  71.7 &  100.0 &  90.4  &   83.9   & 68.9  &  95.5 &  50.2 &   84.2  &  93.3    &   94.9     &   82.2  \\
% \multicolumn{1}{c|}{} & Pixel  \\
\multicolumn{1}{c|}{} & PRO   &  86.4 & 67.9 & 63.9 &  79.4 &  57.8  &  83.9  &  94.8 & 87.1 &   83.6  &  87.5     &   88.4   & 59.4  &56.8 &   79.0   &  73.5 &   76.6  \\ 
\midrule
\multicolumn{1}{r|}{\multirow{3}{*}
{{DiffusionAD~\cite{zhang2023diffusionad}}}}  & Image &  98.1 &  71.4  & 72.1 & 98.6 & 79.8    &  92.2 &  100.0 &  73.6  &   87.6   & 66.1  &  93.2 &  51.0 &   84.7  &  99.6   &   98.5   &   84.4 \\
% \multicolumn{1}{c|}{} & Pixel  \\
\multicolumn{1}{c|}{} & PRO   &  83.7 & 66.8 & 76.5 &  90.9 &  55.9  &  87.4  &  99.3 & 82.2 &   87.8  &  77.9     &   72.5  & 65.8  &51.8 &   87.9   &  94.9 &   78.8  \\

\midrule
\multicolumn{1}{r|}{\multirow{3}{*}{{PaDim~\cite{defard2020padim}}}}  & Image &  99.0 &  57.6  &  71.9 & 91.0  &  55.3    &  93.3 &  97.2  &  68.9  &   53.7   &  49.8  &  84.1 &  79.3 &   84.9  &  97.4    &   78.0     &   77.4  \\
\multicolumn{1}{c|}{} & PRO   &  96.5 & 61.0 & 92.0 &  93.7 &  48.0  &  92.3  &  98.1 & 75.5 &   86.4   &  79.7     &   75.6   & 82.0  &77.1 &   93.8   &  87.6 &   82.6  \\

\midrule
\multicolumn{1}{r|}{\multirow{3}{*}{{SPADE~\cite{spade}}}}   & Image  & 98.3 &70.3  & 76.7 &  85.6  &  43.5 & 86.5  &90.5  &   76.0      &    68.3       &   53.4   &   93.3    &   56.3   &    81.1        &     94.2       &   87.1  &   77.4      \\
\multicolumn{1}{r|}{}                                        & PRO  &  94.5 &  65.3  &  92.3 &  88.0&  70.2 & 94.3 &  97.8 &   79.9   &    88.7   &  84.8    &   74.2&   78.5  &   60.8   &     91.5   &  85.3 &83.1       \\ 
\midrule
\multicolumn{1}{r|}{\multirow{3}{*}{\begin{tabular}[c]{@{}l@{}}PatchCore ~\cite{roth2021towards}\end{tabular}}} & Image  &  100.0      &    77.6   &    78.4     &   98.5    &  69.9  &     96.3    &     99.8      &   75.7   &   80.2  &   50.3   &    97.3  &   77.3   &   88.7 &    99.0  &   92.2     &   85.4      \\
\multicolumn{1}{l|}{}                                        & PRO    &   97.5   &   79.5  & 90.9  &   97.3     &   58.1   &   87.1   &  98.3 &  86.4   &   86.5   &   73.6   &91.2 &    79.8   &    85.8  &   93.3   &  94.4     &   86.7      \\ 

\midrule
\multicolumn{1}{r|}{\multirow{3}{*}{Ours}} & Image  &  99.6 & 65.4  &  80.8 & 78.4 &  48.1 & 97.5  &93.2& 83.0  & 69.7 & 57.5 &86.7  & 69.5  &  66.6 & 97.2  &  79.3 &  78.2 \\
\multicolumn{1}{l|}{}                                       & PRO & 96.1  & 69.0 &  92.4 &  88.3 & 59.7 & 93.7  &  97.1     &  77.7 &  88.1  &  85.8 &  73.8  &  78.8  &   62.2         &  92.0  & 86.7 &  82.8\\ 
\midrule
\multicolumn{18}{c}{Online Methods}\\

\midrule
\multicolumn{1}{r|}{\multirow{3}{*}
{{DRAEM~\cite{zavrtanik2021draem}}}}  & Image &  94.9 &  68.1  &  73.7 & 92.3  &  87.3    &  71.7 &  99.0 &  91.7  &   88.7   & 72.1  &  96.5 &  54.5 &   92.5  &  91.1    &   92.8     &   86.2  \\
% \multicolumn{1}{c|}{} & Pixel  \\
\multicolumn{1}{c|}{} & PRO   &  77.6 & 61.3 & 77.3 &  80.5 &  91.1  &  90.1  &  94.3 & 85.2 &   85.4  &  81.2     &   94.6   & 59.9  &65.8 &   77.6   &  62.3 &   78.9  \\ 

\midrule

\multicolumn{1}{r|}{\multirow{3}{*}
{{DiffusionAD~\cite{zhang2023diffusionad}}}}  & Image &  99.1 &  64.9  &  75.1 & 93.6  &  96.2    &  94.3 &  99.2 &  82.1  &   90.8   & 58.2  &  92.1 &  87.0 &   78.8  &  96.8    &   93.1     &   86.8 \\
% \multicolumn{1}{c|}{} & Pixel  \\
\multicolumn{1}{c|}{} & PRO   &  84.8 & 64.9 & 81.9 &  84.4 &  81.0  &  91.1  &  95.5 & 85.2 &   71.7  &  88.9     &   82.4  & 76.3  &84.0 &   62.6   &  76.8 &   80.0  \\ 

\midrule

\multicolumn{1}{r|}{\multirow{3}{*}
{{PaDim~\cite{defard2020padim}}}}  & Image &  99.0 &  58.0  &  72.7 & 92.6  &  58.2    &  93.2 &  97.3  &  68.4  &   52.7   &  48.8  &  85.4 &  79.3 &   84.5  &  96.8    &   79.0     &   77.7  \\
\multicolumn{1}{c|}{} & PRO   &  96.5 & 61.6 & 92.0 &  94.2 &  50.6  &  92.4  &  98.2 & 76.1 &   86.6   &  79.8     &   76.0   & 82.5  &78.1 &   94.0   &  88.1 &   83.1  \\ 

\midrule
\multicolumn{1}{r|}{\multirow{3}{*}{{SPADE~\cite{spade}}}}   & Image  & 98.1 &77.7  & 78.5 &  90.0  &  49.0 & 93.6  &93.4  &   82.2      &    71.5       &   60.2   &   96.2    &   50.2   &    75.7        &     97.7       &   90.0  &   80.3      \\
\multicolumn{1}{r|}{}                                        & PRO  &  94.6 &  67.3  &  93.8 &  88.5&  79.3 & 95.3 &  97.5 &   88.3   &    89.4   &  90.6    &   71.0&   78.8  &   65.6   &     91.7   &  85.1 &85.1       \\ 
\midrule
\multicolumn{1}{r|}{\multirow{3}{*}{\begin{tabular}[c]{@{}l@{}}PatchCore ~\cite{roth2021towards}\end{tabular}}} & Image  &  97.3      &    77.4   &    81.9     &   95.9    &  67.8  &     95.8    &     95.6      &   71.4   &   83.9  &   53.5   &    94.6  &   78.9   &   89.5 &    98.0  &   91.3     &   84.9      \\
\multicolumn{1}{l|}{}                                        & PRO    &   94.1   &   80.6  & 90.5  &   95.7     &   59.8   &   85.7   &  95.0 &  81.4   &   84.9   &   71.2   &85.7 &    91.4   &    90.5  &   89.5   &  90.6     &   85.8      \\ 

\midrule
\multicolumn{1}{r|}{\multirow{3}{*}{{Ours}}}                 & Image  &  99.8  &  87.0  &  84.4  & 94.2&  59.4 & 97.8 & 98.4 & 96.8 &  80.9 & 60.8 &  88.6 &  89.9 & 82.8  & 99.0     &  91.2 &  \bf  87.4     \\
\multicolumn{1}{c|}{}                                        & PRO    & 96.5  &  76.2 & 93.5 & 95.5& 81.2  & 92.7& 98.4 &  82.8 & 90.3 & 87.3 & 78.5 &  87.6 &  72.9 &  94.6  &  92.9      &   \bf 88.1      \\
\bottomrule
\end{tabular}
}
\label{detail_newone}
\end{sidewaystable}

% \appendices
\subsection{Proof of the Equation (8) in Section 3.3}
\label{proof}
% \begin{proof}
If the original vectors assigned to $\mathcal{A}_i$ are $X^o$, newly assigned are $X^c$. The total vectors are $X$ and its mean is $\mu$, then:
\footnotesize
\begin{equation*}
\begin{split}
    \Sigma_{\mathcal{A}_i}&=\frac{1}{N+N_{\mathcal{A}_i}-1}\sum_{m=1}^{N+N_{\mathcal{A}_i}}(X_m-\mu)(X_m-\mu)^T\\
   & =\frac{1}{N+N_{\mathcal{A}_i}-1}(\sum_{m=1}^{N_{\mathcal{A}_i}}((X_m-\mu^o)-(\mu-\mu^o))((X_m-\mu^o)\\
   &  -(\mu-\mu^o))^T+\sum_{m=n+1}^{n+m}...)\\
   &=\frac{1}{N+N_{\mathcal{A}_i}-1}(\sum_{m=1}^{N_{\mathcal{A}_i}}(X_m-\mu^o))(X_m-\mu^o)^T- \sum_{m=1}^{N_{\mathcal{A}_i}}\\
   &  (X_m-\mu^o)(\mu-\mu^o)^T-\sum_{m=1}^{N_{\mathcal{A}_i}}(\mu-\mu^o)(X_m-\mu^o)^T+\\
   &\sum_{m=1}^{N_{\mathcal{A}_i}}(\mu-\mu^o)(\mu-\mu^o)^T+\sum_{m=n+1}^{n+m}...)\\
   &=\frac{1}{N+N_{\mathcal{A}_i}-1}((N_{\mathcal{A}_i}-1)\Sigma^o-(\sum_{m=1}^{N_{\mathcal{A}_i}}(X_m-\mu^o))(\mu-\mu^o)^T\\
   &  -(\mu-\mu^o)\sum_{m=1}^{N_{\mathcal{A}_i}}(X_m-\mu^o)^T+\sum_{m=1}^{N_{\mathcal{A}_i}}(\mu-\mu^o)(\mu-\mu^o)^T\\
   &  +\sum_{m=n+1}^{n+m}...)\\
   &=\frac{1}{N+N_{\mathcal{A}_i}-1}((N_{\mathcal{A}_i}-1)\Sigma^o+N_{\mathcal{A}_i}(\mu-\mu^o)(\mu-\mu^o)^T \\
   &  +(N-1)\Sigma^c+N(\mu-\mu^c)(\mu-\mu^c)^T).
\end{split}
\end{equation*}
% \end{proof}

\section*{Acknowledgements}
This work was supported by National Key R\&D Program of China under Grant No. 2021ZD0110400, the Fundamental Research Funds for the Central Universities No.xxj032023020.and sponsored by the CAAI-MindSpore OpenFund, developed on Openl Community.



%% References


\bibliographystyle{model1-num-names}
\bibliography{ref}






% \end{thebibliography}

\end{sloppypar}
\end{document}